\documentclass{article}

\PassOptionsToPackage{numbers, compress}{natbib}

\usepackage[preprint]{neurips_2025}

\usepackage[utf8]{inputenc} 
\usepackage[T1]{fontenc}    
\usepackage{hyperref}       
\usepackage{booktabs}       
\usepackage{amsfonts}       
\usepackage{nicefrac}       
\usepackage{microtype}      
\usepackage{xcolor}         
\usepackage{url}
\usepackage{amsmath}
\usepackage{xcolor}
\usepackage{graphicx}
\usepackage{multirow, makecell}
\usepackage{bbding}
\usepackage{pifont}
\usepackage{wasysym}
\usepackage{amssymb}
\usepackage{caption}
\usepackage{subcaption}
\usepackage{wrapfig}
\usepackage{tocloft}
\usepackage{etoc}
\usepackage{lipsum}

\newcommand{\ie}{\textit{i}.\textit{e}.}
\newcommand{\eg}{\textit{e}.\textit{g}.}

\newcommand{\vs}{\textit{vs}.\ }

\definecolor{ignorecolor}{rgb}{0.875,0.875,0.75}

\usepackage[normalem]{ulem}
\useunder{\uline}{\ul}{}

\usepackage{mathtools}

\usepackage{algorithm}
\usepackage{algpseudocode}
\usepackage{verbatim}

\title{SViMo: Synchronized Diffusion for Video and Motion Generation in Hand-object Interaction Scenarios}

\author{%
    Lingwei Dang\footnotemark[1]\textsuperscript{\hspace{0.6em}\rm 1},
    Ruizhi Shao\thanks{Equal contributions. Email: levondang@163.com, jia1saurus@gmail.com.}\textsuperscript{\hspace{0.6em}\rm 2},
    Hongwen Zhang\textsuperscript{\hspace{0.6em}\rm 3},\\
    \textbf{Wei MIN}\textsuperscript{\hspace{0.6em}\rm 4}\textbf{,}
    \textbf{Yebin Liu}\footnotemark[2]\textsuperscript{\hspace{0.6em}\rm 2}\textbf{,}
    \textbf{Qingyao Wu}\thanks{Corresponding Authors. Email: liuyebin@mail.tsinghua.edu.cn, qyw@scut.edu.cn.}\textsuperscript{\hspace{0.6em}\rm 1}\\
  \textsuperscript{\rm 1}School of Software Engineering, South China University of Technology \\
  \textsuperscript{\rm 2}Department of Automation, Tsinghua University \\
  \textsuperscript{\rm 3}School of Artificial Intelligence, Beijing Normal University \\
  \textsuperscript{\rm 4}Shadow AI
}

\begin{document}

\maketitle


\begin{abstract}
    Hand-Object Interaction (HOI) generation has significant application potential. However, current 3D HOI motion generation approaches heavily rely on predefined 3D object models and lab-captured motion data, limiting generalization capabilities. Meanwhile, HOI video generation methods prioritize pixel-level visual fidelity, often sacrificing physical plausibility. Recognizing that visual appearance and motion patterns share fundamental physical laws in the real world, we propose a novel framework that combines visual priors and dynamic constraints within a synchronized diffusion process to generate the HOI video and motion simultaneously. To integrate the heterogeneous semantics, appearance, and motion features, our method implements tri-modal adaptive modulation for feature aligning, coupled with 3D full-attention for modeling inter- and intra-modal dependencies. Furthermore, we introduce a vision-aware 3D interaction diffusion model that generates explicit 3D interaction sequences directly from the synchronized diffusion outputs, then feeds them back to establish a closed-loop feedback cycle. This architecture eliminates dependencies on predefined object models or explicit pose guidance while significantly enhancing video-motion consistency. Experimental results demonstrate our method's superiority over state-of-the-art approaches in generating high-fidelity, dynamically plausible HOI sequences, with notable generalization capabilities in unseen real-world scenarios. Project page at \href{https://droliven.github.io/SViMo_project/}{https://droliven.github.io/SViMo\_project/}.
\end{abstract}
\section{Introduction}
\label{sec:introduction}

\begin{figure}[!t]
    \centering
    \includegraphics[width=1.0\linewidth]{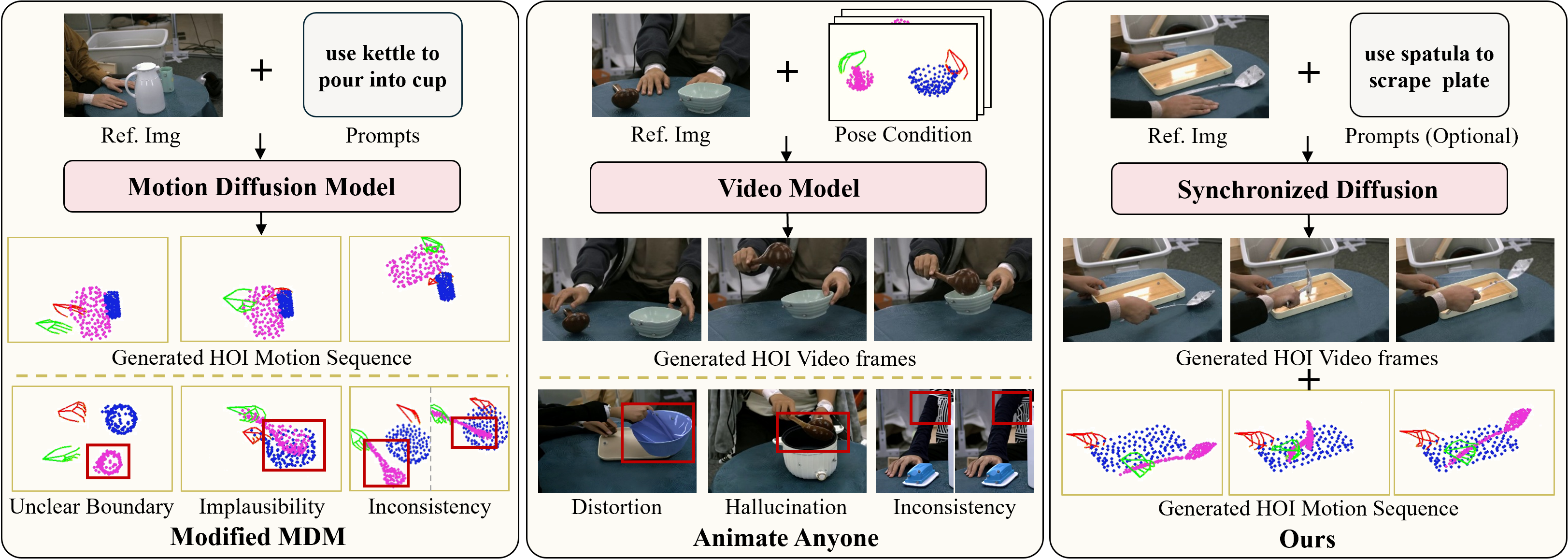}
    \caption{Different HOI generation methods. 
    Approaches like MDM~\cite{tevet2023human} rely on limited mocap data without visual guidance, resulting in blurred boundaries and compromised plausibility and consistency.
    Methods like Animate Anyone~\cite{hu2024animate} leverage large-scale visual priors but exhibit distortions and inconsistencies because of inadequate physical awareness.
    Our method marries visual priors with 3D motion constraints and eliminates dependency on pre-defined object models or pose guidance.}
    \label{fig:teaser}
\end{figure}

Human-object or hand-object interaction (HOI) generation serves critical applications across gaming, animation, digital human creation, and robotic action retargeting~\cite{gao2024coohoi,liu2024geneoh,2024Hand,wang2023deepsimho}.
Some studies~\cite{cha2024text2hoi,diller2024cg,zhang2024manidext} construct high-precision 3D interaction datasets through laboratory-based multi-view camera arrays and motion capture systems, then train diffusion-based motion generators. 
These object-centric approaches typically predict parametric human/hand motions and the corresponding object pose sequences given object meshes and initial configurations. 
However, existing datasets~\cite{liu2022hoi4d,zhan2024oakink2,liu2024taco} collected in laboratory environments lack diversity in object types and interaction patterns, constraining model generalization and resulting in ambiguous object boundaries, implausible or inconsistent actions (Fig.~\ref{fig:teaser}, left). 
Moreover, the reliance on precise 3D object models fundamentally limits their zero-shot generation capabilities.

Recent advances in large video foundation models based on Diffusion Transformers (DiT)~\cite{peebles2023scalable} (\eg, Sora~\cite{brooks2024video}, CogVideo~\cite{hong2023cogvideo,yang2025cogvideox}, HunyuanVideo~\cite{kong2024hunyuanvideo}), have shown impressive capabilities in modeling physical dynamics through large-scale video training. 
These models can generate interaction videos with high visual fidelity end-to-end from text or reference images. 
However, their pixel-level generation approaches often struggle to produce accurate and coherent hand-object interactions due to limited explicit modeling of motion dynamics and physical constraints.
To address this, some methods extend image-based diffusion models (\eg, SVD~\cite{blattmann2023stable}) by adding pose-guided pipelines~\cite{xu2024anchorcrafter,hu2024animate,zhu2024champ,xu2024magicanimate}. 
These approaches combine pose conditions and appearance features to improve human-object interaction generation. 
While effective, they require pose sequences or externally estimated motion trajectories as inputs, preventing full end-to-end text/image-conditioned generation. 
Additionally, their single-frame generation leads to poor temporal coherence, causing flickering and identity inconsistencies (Fig.~\ref{fig:teaser}, middle).



These challenges reveal a longstanding methodological contradiction: motion generation systems excel at physical constraint modeling but suffer from limited data scales. In contrast, video generation models leverage massive visual priors but lack motion plausibility. 
We argue that this division stems from neglecting the co-evolution mechanism between visual appearance and motion patterns: they not only share the same foundation of physical dynamics but also could leverage similar diffusion processes. 
Based on this insight, we propose SViMo, a \textbf{S}ynchronized \textbf{Vi}deo-\textbf{Mo}tion diffusion framework that enables synchronous HOI video generation and motion synthesis within a unified architecture (Fig.~\ref{fig:teaser}, right).
The core innovation lies in extending a pretrained image-to-video foundation model into the multimodal joint generation framework through the scalable DiT architecture.
To better integrate the heterogeneous features of text semantics, visual appearance, and motion dynamics, we introduce the triple modality adaptive modulation to align feature scales and employ a 3D full-attention mechanism to learn their synergistic and complementary dependencies.
Additionally, it is difficult for video foundation models to learn explicit 3D interaction motions directly. To bridge the representation gap and reduce optimization complexity, we project 3D motions onto 2D image planes, constructing ``rendered motion videos'' as SViMo's motion representation.

To further enhance the video-motion consistency, we design a Vision-aware 3D Interaction Diffusion model (VID). This model generates explicit 3D hand poses and object point clouds using denoised latent codes from the synchronized diffusion, which are then reinjected into the SViMo as interaction guidance and gradient constraints.
Unlike methods requiring pre-specified action series, our approach integrates video synthesis and 3D interaction generation within an end-to-end denoising pipeline. This creates a closed-loop feedback mechanism where motion guidances refine video generation while video latents update motion results, enabling synergistic co-evolution of both modalities.

In summary, our contributions are threefold:
\begin{itemize}  
\item A novel synchronized diffusion model for joint HOI video and motion denoising, effectively integrating large-scale visual priors with motion dynamic constraints.  
\item A vision-aware 3D interaction diffusion that generates explicit 3D interaction sequences, forming a closed-loop optimization pipeline and enhancing video-motion consistency.
\item Our method generates HOI video and motion synchronously without requiring pre-defined poses or object models. Experiment results demonstrate superior visual quality, motion plausibility, and generalization capability to unseen real-world data.
\end{itemize}

\section{Related Work}


\textbf{3D interaction synthesis} relies on high-precision motion capture datasets, some of which focus on human action conditioned on static objects~\cite{hassan2021stochastic,zhang2022couch}, while others simultaneously capture interactions of both human and dynamic objects~\cite{chao2021dexycb,liu2022hoi4d,yang2022oakink,zhan2024oakink2,liu2024taco,taheri2020grab,fan2023arctic}. Building upon these datasets, existing 3D interaction generation methods employ diffusion models to either introduce intermediate contact maps or milestones for modeling kinematic features~\cite{cha2024text2hoi,diller2024cg,li2024controllable,pi2023hierarchical,peng2023hoi,zhang2024manidext,xu2023interdiff,lee2024interhandgen,kulkarni2024nifty,li2023object,liu2024primitive,li2024task,liu2024geneoh} or leverage physical simulations to ensure physical dynamics plausibility~\cite{xu2024interdreamer,wang2023physhoi,braun2024physically,xu2025intermimic,luo2024omnigrasp}. 
However, due to the limited availability of 3D interaction data, the generalization capability of these approaches remains constrained. In contrast, our method leverages large-scale visual priors, operates conveniently without requiring 3D object models, and demonstrates promising generalization potential.


\textbf{Interaction Video Generation.} The success of image generation models~\cite{ho2020denoising,ramesh2021zero,rombach2022high,saharia2022photorealistic} has inspired video generation approaches. Some works~\cite{blattmann2023stable,xu2024anchorcrafter,pang2024manivideo,hu2024animate,zhu2024champ,xu2024magicanimate,xu2024easyanimate} extend 2D image denoising U-Nets into video models by inserting temporal attention layers, and enhance controllability through pose guider and reference net. Other native large video models~\cite{brooks2024video,hong2023cogvideo,yang2025cogvideox,kong2024hunyuanvideo,jiang2025vace} directly generate videos based on 3D DiT~\cite{peebles2023scalable} networks. While the results of these methods are visually realistic, they lack awareness of 3D physical dynamics.
Contrastively, our method enhances the dynamic plausibility of generated videos by introducing dynamic information through synergistic motion denoising and a 3D interaction diffusion model.

\textbf{Multimodal Generative Models}. Driven by advancements in vision-language models~\cite{chen2024internvl,liu2023visual,wang2024cogvlm,bai2025qwen2,yang2023dawn}, researchers have explored versatile generative models that align with the multimodal essence of the physical world.
Some works~\cite{jeong2023power,yariv2024diverse} cascade single-modality generators into asynchronous pipelines, yet suffer from complex workflows and noise accumulation. 
Others develop end-to-end multimodal joint generation frameworks that rely on massive aligned multimodal data~\cite{ao2024body,li2024visionlanguage,guo2024prediction,zitkovich2023rt}, indirect bridging mechanisms~\cite{tang2023any}, or intricate hierarchical attention strategy~\cite{liu2025javisdit} to ensure cross-modal synchronization.
Differently, our method extends a native large video model into synchronized video-motion generation systems, aligning heterogeneous modality features through multimodal adaptive modulation and closed-loop feedback strategies.

\section{Methodology}
\label{sec:method}


We define the HOI video and motion generation task as follows: Given a reference image frame $\boldsymbol{I}\in \mathbb{R}^{H \times W \times 3}$ and a textual prompt $\boldsymbol{P}$, generate the future video $\boldsymbol{V} \in \mathbb{R}^{N \times H \times W \times 3}$ and 3D motion sequence $\boldsymbol{M}=\{(\boldsymbol{h}_i, \boldsymbol{o}_i)\}_{i=1}^{N}$ with $N$ time steps. 
Here $\boldsymbol{h}_i \in \mathbb{R}^{J \times 3}$ and $\boldsymbol{o}_i \in \mathbb{R}^{K \times 3}$ are the hand joints trajectories and object point clouds, each has $J$ and $K$ nodes respectively.
The following sections detail each component of our approach. See the \textbf{Appendix} for training/inference pseudo-code and additional details.

\begin{figure}[!t]
    \centering
    \includegraphics[width=1.0\linewidth]{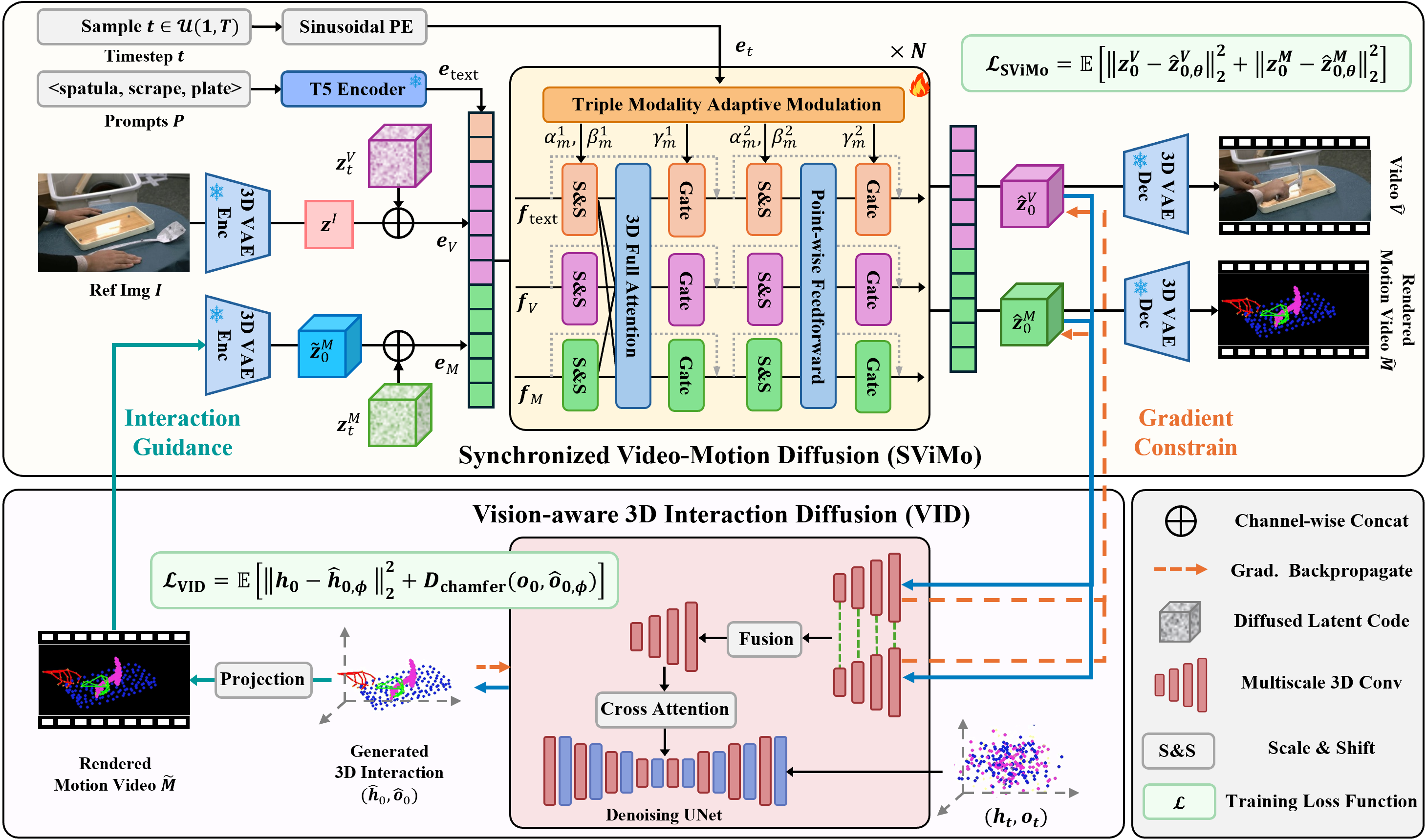}
    \caption{Our method comprises: (1) A synchronous diffusion model that jointly generates HOI videos and motions (Sec.~\ref{sec:svimo}). (2) A vision-aware interaction diffusion model that generates 3D hand pose trajectories and point clouds from the former's outputs (Sec.~\ref{sec:vid}), then feeds them back into the synchronized denoising process to establish closed-loop optimization (Sec.~\ref{sec:training}).}
    \label{fig:pipelline}
\end{figure}

\subsection{Preliminary: Basic Large Video Generation Model}
\label{sec:preliminary}


The core of large video generation models is diffusion and its reverse denoising process~\cite{ho2020denoising}. The diffusion process adds Gaussian noise to video latent codes $\boldsymbol{z}$ step-by-step using $q(\boldsymbol{z}_t | \boldsymbol{z}_{t-1}) = \mathcal{N}(\boldsymbol{z}_t; \sqrt{\alpha_t} \boldsymbol{z}_{t-1}, \beta_t \boldsymbol{I})$, where $\beta_t \in [0,1]$ increases monotonically with time step $t$, and $\alpha_t = 1-\beta_t$. Repeated application of this formula yields the marginal distribution:
\begin{equation}
    \label{eq:forward_diffusion}
    q(\boldsymbol{z}_t | \boldsymbol{z}_{0}) = \mathcal{N}(\boldsymbol{z}_t; \sqrt{\bar{\alpha}_t} \boldsymbol{z}_0, (1 - \bar{\alpha}_t) \boldsymbol{I}), \quad \bar{\alpha}_t = \prod_{1}^{t} (1-\beta_t).
\end{equation}
The above distribution converges to an isotropic Normal distribution when $t$ grows large.

Correspondingly, the reverse denoising generator $\mathcal{G}$  with trainable weights $\theta$ predicts the clean latent code $\hat{\boldsymbol{z}}_{0,\theta}$, and is optimized by: 
\begin{equation}
    \label{eq:x0_prediction}
    \mathcal{L}_{\text{x}_0 \text{-prediction}} = \mathbb{E}_{\boldsymbol{z}, \boldsymbol{c}, t, \boldsymbol{\epsilon} \sim \mathcal{N}(0,\boldsymbol{I})} \left[ \| \boldsymbol{z}_0 - \mathcal{G}_\theta (\boldsymbol{z}_t, \boldsymbol{c}, t) \|_2^2 \right].
\end{equation}
During the inference phase, the trained model is utilized to iteratively denoise from pure Gaussian noise step by step, following the procedure:
\begin{equation}
    \label{eq:reverse_diffusion}
    \left\{\begin{aligned}
        \hat{\boldsymbol{z}}_{0, \theta} &= \mathcal{G}_\theta(\boldsymbol{z}_t, \boldsymbol{c}, t), \\
        \boldsymbol{z}_{t-1} &\sim \mathcal{N}\left(\frac{\sqrt{\alpha_t} (1-\bar{\alpha}_{t-1})}{1-\bar{\alpha}_{t}} \boldsymbol{z}_t + \frac{\sqrt{\bar{\alpha}_{t-1}} (1-\alpha_t)}{1-\bar{\alpha}_{t}} \hat{\boldsymbol{z}}_{0, \theta}, \frac{1 - \bar{\alpha}_{t-1}}{1 - \bar{\alpha}_t} (1 - \alpha_t) \boldsymbol{I}\right).
    \end{aligned}
    \right.
\end{equation}
Finally, we obtain the clean denoised latent code $\hat{\boldsymbol{z}}_{0}^{\star}$, which is then decoded to the raw video space.

\subsection{Framework Overview}
\label{sec:overview}

Based on the insight that visual appearance and motion dynamics share inherent physical laws of the real world, we propose an end-to-end framework that unifies HOI video and 3D motion generation by integrating visual priors with dynamic constraints. 
Our framework consists of two key components. The first is SViMo, which focuses on video generation (Fig.~\ref{fig:pipelline}, top; Sec.~\ref{sec:svimo}). 
It extends a pre-trained image-to-video foundation model into a joint video-motion generation architecture. During joint denoising, SViMo dynamically aligns visual appearance with high-level motion signals, improving both visual quality and motion plausibility. 
Notably, the motion representations in SViMo refer not to explicit 3D hand trajectories or object point cloud sequences.
Because there is a significant gap between 3D motion data and 2D video representations, directly modeling explicit 3D motions would disrupt the pre-trained foundation model's visual priors and degrade performance. 
Instead, we project explicit 3D interactions onto the 2D image plane to create ``rendered motion videos'' as intermediate representations for SViMo. 
The second component is VID, which generates explicit 3D motions (Fig.~\ref{fig:pipelline}, bottom; Sec.~\ref{sec:vid}). It maps the output of SViMo, both the generated video and the 2D rendered motion video, into target 3D interactions, \ie, hand-object trajectories and object point clouds. 
Additionally, the collaboration between SViMo and VID forms a closed-loop feedback pipeline, ensuring consistency between generated videos and 3D motions (Sec.~\ref{sec:training}).

\subsection{Synchronized Video-Motion Diffusion}
\label{sec:svimo}

The SViMo learns to predict the video and motion, given the time step, text prompt, and reference image: $(\hat{\boldsymbol{z}}_0^V, \hat{\boldsymbol{z}}_0^M) = \mathcal{G}_\theta \left((\boldsymbol{z}_t^V, \boldsymbol{z}_t^M), (\boldsymbol{P}, \boldsymbol{I}), t \right)$. 
The entire video-motion generation process comprises feature embedding, multimodal feature modulation and fusion, and joint denoising.

\textbf{Feature Embedding}.
\label{sec:emb}
Time step $t$ is sampled from a uniform distribution and then added sinusoidal position encoding, followed by a simple two-layer linear mapping to obtain the time embedding $\boldsymbol{e}_t \in \mathbb{R}^{d_\text{time}}$. 
For the text prompt, we use a frozen pre-trained language model, Google T5~\cite{raffel2020exploring}, to extract text embedding $\boldsymbol{e}_\text{text} \in \mathbb{R}^{L \times d_p}$, and then calculate the semantics features $\boldsymbol{f}_\text{text}$ with linear projection, where $L$ is the max length of textual tokens. 
Thirdly, for the original reference image $\boldsymbol{I}$, to ensure a certain level of robustness, we first add random noise yielding $\boldsymbol{I}_\text{noised}$. Then, we compute the compressed latent code $\boldsymbol{z}_I$ through a 3D video VAE, $\boldsymbol{z}_I = \mathcal{E}(\boldsymbol{I}_\text{noised}) \in \mathbb{R}^{\frac{H}{rh} \times \frac{W}{rw} \times d_\text{VAE}}$. 
Additionally, for the target video $\boldsymbol{V}$, we also map it to the latent space with the same VAE. Then the noised latent code $\boldsymbol{z}_{t}^{V} \in \mathbb{R}^{(\frac{N-1}{rn} +1) \times \frac{H}{rh} \times \frac{W}{rw} \times  d_\text{VAE}}$ are obtained based on the forward diffusion process in Eq.~\ref{eq:forward_diffusion}. 
To better align it with the image condition signal during the video denoising process, we repeat the image feature $\boldsymbol{z}_I$ along the temporal axis, and concatenate them along the channel-axis to get video feature embedding $\boldsymbol{e}_V \in \mathbb{R}^{(\frac{N-1}{rn} +1) \times \frac{H}{rh} \times \frac{W}{rw} \times  (2 \times d_\text{VAE})} = \boldsymbol{z}_t^V \oplus \boldsymbol{z}_I$. Then the visual feature is patchified through a 2D stride convolution, esulting in $\boldsymbol{f}_\text{V} \in \mathbb{R}^{\left[ (\frac{N-1}{rn} +1) \times \frac{H}{2 \times rh} \times \frac{W}{2 \times rw}  \right] \times  d}$. 
The rendered motion video is encoded through the same VAE and diffused to obtain $\boldsymbol{z}_t^M$. 
In contrast, to get motion embedding $\boldsymbol{e}_M$, the diffused latent code here are concatenated with interaction guidance $\tilde{\boldsymbol{z}}_0^M$ provided by VID (Sec.~\ref{sec:vid}), rather than being combined with $\boldsymbol{z}_I$. 
After that, the motion feature $\boldsymbol{f}_M$ is obtained in the same way as that of $\boldsymbol{f}_V$.

\textbf{Multimodal Feature Modulation and Fusion.}
\label{sec:modulation}
In our SViMo framework, the DiT token sequence comprises three distinct modalities: text tokens $\boldsymbol{f}_\text{text}$, video tokens $\boldsymbol{f}_V$ and motion tokens $\boldsymbol{f}_M$, which differ significantly in both feature spaces and numerical scales. 
To bridge these disparities while preserving modality-specific characteristics, we adopt a triple modality adaptive modulation method that learns modulation parameters from the timestep signal $\boldsymbol{e}_t$. These parameters determine the scaling, shifting, and gating operations of each modality's features separately.
Additionally, a 3D full-attention mechanism is employed to capture intra- and inter-modal relationships.
Take the processing of text features as an example, the DiT Block $\mathcal{B}$ proceeds as follows:
\begin{equation}
    \label{eq:three_expert}
    \mathcal{B}(\cdot) = \left\{
    \begin{aligned}
        \{\boldsymbol{\alpha}_\text{text}^i, \boldsymbol{\beta}_\text{text}^i, \boldsymbol{\gamma}_\text{text}^i\}_{i=1}^{2} &= \text{MLP}(\boldsymbol{e}_t),  \\
        \boldsymbol{f}_\text{text}^{\prime} = \text{LN}(\boldsymbol{f}_\text{text}) &\odot (1+\boldsymbol{\alpha}_\text{text}^1) + \boldsymbol{\beta}_\text{text}^1, \\
        \boldsymbol{f}_\text{text}^{\prime} = \boldsymbol{f}_\text{text}^{\prime} + \boldsymbol{\gamma}_\text{text}^1 &\odot  \cap_\text{text} \left( \text{3DFA} \left(  \cup (\boldsymbol{f}_\text{text}^{\prime},\boldsymbol{f}_V^{\prime},\boldsymbol{f}_M^{\prime}) \right) \right), \\
        \boldsymbol{f}_\text{text}^{\prime \prime} = \text{LN}(\boldsymbol{f}_\text{text}^{\prime}) &\odot (1+\boldsymbol{\alpha}_\text{text}^2) + \boldsymbol{\beta}_\text{text}^2, \\
    \boldsymbol{f}_\text{text}^{\prime \prime} = \boldsymbol{f}_\text{text}^{\prime \prime}  + \boldsymbol{\gamma}_\text{text}^2 &\odot  \cap_\text{text} \left( \text{FFD} \left(  \cup(\boldsymbol{f}_\text{text}^{\prime},\boldsymbol{f}_V^{\prime},\boldsymbol{f}_M^{\prime})\right) \right),
    \end{aligned}
    \right\} \text{same for } \boldsymbol{f}_V \text{ and } \boldsymbol{f}_M,
\end{equation}
where ``$\text{3DFA}$'' and ``$\text{FFD}$'' are unified multi-head 3D full-attention and feedforward layers, $\cup$ and $\cap$ denote token concatenation and segmentation along the sequence dimension, respectively.

\textbf{Video-Motion Joint Denoising.} The network consists of $N$ stacked DiT blocks. The video and motion features output by the final DiT block are then go through an MLP to reconstruct the VAE latent codes, yielding $\hat{\boldsymbol{z}}_0^V$ for video and $\hat{\boldsymbol{z}}_0^M$ for motion. Finally, the SViMo is optimized according to Eq.~\ref{eq:x0_prediction}:
\begin{equation}
    \label{eq:svimi_loss}
    \mathcal{L}_{\text{SViMo}} = \mathbb{E}_{(\boldsymbol{z}^V, \boldsymbol{z}^M), (\boldsymbol{P}, \boldsymbol{I}), t,\boldsymbol{\epsilon}} \left[ \| \boldsymbol{z}_0^V - \mathcal{G}_\theta(\boldsymbol{z}_t^V, (\boldsymbol{P}, \boldsymbol{I}), t) \|_2^2 + \| \boldsymbol{z}_0^M - \mathcal{G}_\theta(\boldsymbol{z}_t^M, (\boldsymbol{P}, \boldsymbol{I}), t) \|_2^2\right].
\end{equation}

\subsection{Vision-aware 3D Interaction Diffusion}
\label{sec:vid}

The VID $\mathcal{M}_\phi$ generates the explicit 3D hand pose trajectories $\hat{\boldsymbol{h}}_0$ and object point cloud sequences $\hat{\boldsymbol{o}}_0$  given latent codes of videos $\boldsymbol{z}_t^V$ and motions (rendered motion video) $\boldsymbol{z}_t^M$ at any time. The framework operates as follows:
First, a dual-stream 3D convolutional module encodes multi-scale spatiotemporal features from both video and motion codes. Then they are fused and subsequently injected into the 3D interaction denoising modules through cross-attention mechanisms to synthesize 3D HOI trajectory sequences.
Following the $\text{x}_0$-prediction formula in Eq.~\ref{eq:x0_prediction}, the model is optimized through the following loss function:
\begin{equation}
    \label{eq:vid_loss}
    \left\{\begin{aligned}
        (\hat{\boldsymbol{h}}_{0, \phi}, \hat{\boldsymbol{o}}_{0, \phi}) &= \mathcal{M}_\phi \left[ (\boldsymbol{h}_{t}, \boldsymbol{o}_{t}),({\boldsymbol{z}}_{t}^V, {\boldsymbol{z}}_{t}^M), t \right], \\
        \mathcal{L}_\text{VID} &= \mathbb{E}_{(\boldsymbol{h}, \boldsymbol{o}), (\hat{\boldsymbol{z}}_\theta^V, \hat{\boldsymbol{z}}_\theta^M), t, \boldsymbol{\epsilon}}\left[\text{MSE}(\boldsymbol{h}_0, \hat{\boldsymbol{h}}_{0, \phi}) + \text{D}_{\text{chamfer}}(\boldsymbol{o}_0, \hat{\boldsymbol{o}}_{0, \phi})\right].
    \end{aligned}
    \right.
\end{equation}

\subsection{Close-loop Feedback and Training Objectives}
\label{sec:training}

\textbf{Close-loop Feedback}. To enhance the mutual promotion and co-evolution between SViMo and VID, and improve the consistency between video and 3D motions, we design a closed-loop feedback mechanism. This mechanism includes two pathways: interaction guidance and gradient constraint.
Specifically, for the straightforward interaction guidance strategy (Eq.~\ref{eq:loop}, row 1st), we first generate the 3D interaction from the video and motion inputs of SViMo: $(\tilde{\boldsymbol{h}}_0, \tilde{\boldsymbol{o}}_0) = \mathcal{M}_\text{no-grad} \left((\boldsymbol{h}_t, \boldsymbol{o}_t), (\boldsymbol{z}_t^V, \boldsymbol{z}_t^M), t \right)$, then projected it onto the 2D image plane to obtain rendered motion video $\tilde{\boldsymbol{M}}$, which are subsequently embedded into the VAE latent space yielding $\tilde{\boldsymbol{z}}_0^M$. 
Finally, it is concatenated with the noised motion latent code $\boldsymbol{z}_t^M$ mentioned in Sec.~\ref{sec:emb}, forming an additional interaction guidance for the SViMo.
On the other hand, the input of VID could comes from the output of the SViMo. Therefore, the gradient of VID will be backpropagated into the SViMo during the training process, forming a gradient constraint path and thereby promoting its optimization (Eq.~\ref{eq:loop}, row 2nd).
\begin{equation}
    \label{eq:loop}
    \left\{\begin{aligned}
        &(\boldsymbol{z}_t^V, \boldsymbol{z}_t^M) \xrightarrow{\text{VID}_\text{no-grad}} (\tilde{h}_0, \tilde{o}_0) \xrightarrow{\text{Proj.}} \tilde{\boldsymbol{M}} \xrightarrow{\text{VAE } \mathcal{E}} \tilde{\boldsymbol{z}}_0^M \xrightarrow{\text{Inter. Guid. to SViMo}} (\boldsymbol{z}_t^V, \boldsymbol{z}_t^M \oplus \tilde{\boldsymbol{z}}_0^M), \\
        & (\boldsymbol{z}_t^V, \boldsymbol{z}_t^M \oplus \tilde{\boldsymbol{z}}_0^M) \xleftharpoondown[\text{Gradient}]{\xrightharpoonup[]{\text{SViMo}}} (\hat{\boldsymbol{z}}_0^V, \hat{\boldsymbol{z}}_0^M) \xleftharpoondown[\text{Gradient}]{\xrightharpoonup[]{\ \ \ \text{VID} \ \ \ }} (\hat{\boldsymbol{h}}_0, \hat{\boldsymbol{o}}_0)  \xleftharpoondown[\text{Gradient}]{\xrightharpoonup[]{\text{\ \ Loss\ \ }}} \mathcal{L}_\text{VID}.
    \end{aligned}
    \right.
\end{equation}

\textbf{Training Objectives.} The training process of our method involves two phases: initially warming up the VID based on Eq.~\ref{eq:vid_loss}, followed by closed-loop training where the SViMo and the VID are jointly optimized according to Eq.~\ref{eq:vid_loss} and Eq.~\ref{eq:svimi_loss}: $\mathcal{L} = \omega_1 \mathcal{L}_{\text{SViMo}} + \omega_2 \mathcal{L}_{\text{VID}}$.

\section{Experiments}
\label{sec:exp}


We conducted extensive experiments to validate the effectiveness of our proposed method. More information, such as additional results and limitation discussions, is provided in the \textbf{Appendix}.

\subsection{Experimental Setup}
\label{sec:setup} 

\textbf{TACO Datasets}~\cite{liu2024taco} is a large-scale bimanual hand-object interaction dataset capturing diverse tool-use behaviors via multi-view video recordings and high-fidelity 3D motion annotations. 
Each task is defined as a triplet <tool category, action type, target object category>, describing tool-mediated interactions with objects.
The dataset includes 2.5k interaction sequences, covering 20 object categories, 196 3D models, 14 participants, and 15 daily interaction types. 
It provides allocentric (4096$\times$3000) and egocentric (1920$\times$1080) video streams, totaling 5.2M frames at 30 Hz. For experiments, the dataset is split into training and test sets at a 1:9 ratio.
To reduce computational load, we crop hand-object interaction regions, adjust their aspect ratio to 3:2, and downsample to 49 frames at 8 FPS. This results in videos with spatiotemporal resolution 416$\times$624$\times$49, aligning with CogVideoX's~\cite{yang2025cogvideox} default settings while lowering spatial resolution.

\textbf{Evaluation Metric.} For video evaluation, we use VBench~\cite{huang2024vbench} to assess two key dimensions: Content Quality (including \textbf{Subject Consistency} and \textbf{Background Consistency}) and Dynamic Quality (\textbf{Temporal Smoothness} and \textbf{Dynamic Degree}). To address the partiality of individual metrics, we multiply them to derive a \textbf{Overall} score for holistic evaluation.
For 3D interaction evaluation, we separately assessed hand poses and object point cloud sequences. For the former, we calculated \textbf{MPJPE} (Mean Per Joint Position Error) and \textbf{Motion Smoothness} metrics. For object evaluation, we measured the \textbf{Chamfer Distance} between generated and ground-truth point clouds.
Additionally, we compute a comprehensive \textbf{FID} score via a pretrained interaction autoencoder.

\subsection{Implementation Details}
\label{sec:imp}

\textbf{Network architecture.} 
Our proposed model generates hand-object interaction videos with resolution $[H, W, N] = [416, 624, 49]$ and 3D motion sequences containing $J=42$ hand keypoints and $K=298$ object nodes. 
During training, timesteps are uniformly sampled from [0, 1000], and the embedding dimension is $d_{\text{time}}=512$, while only 50 steps are sampled during inference for acceleration. 
Text prompts (max length $L=226$) are embedded into features of dimension $d_p=4096$. 
The video VAE uses spatial-temporal compression ratios $[rw, rh, rn] = [8, 8, 4]$, producing latent codes of size $[52 \times 78 \times 13]$ with $d_\text{VAE}=16$ channels. 
After patchification via two-step stride convolution, video and motion tokens are compressed to 13182 tokens each. 
This results in a total multimodal token sequence length of 
$226 + 13182 \times 2 = 26590$ (text + video + motion). The DiT backbone comprises 42 DiT Blocks, each with 48 attention heads, totaling 6.02B parameters.

\textbf{Training Details.} All models are trained on 4 NVIDIA A800-80G GPUs. With memory optimization techniques including DeepSpeed ZeRO-3~\cite{rajbhandari2020zero}, gradient checkpointing, and BF16 mixed-precision trick, we achieve a per-GPU batch size of 4. We first warm up the VID for 5k steps, then conduct joint training with the SViMo. To enhance computational efficiency, we initially train at reduced resolution $[H^\prime, W^\prime] = [240, 368]$ for 30k steps before fine-tuning at full resolution for 5k steps. The weights of SViMo and VID terms in the training objectives are $\omega_1 = 1$ and $\omega_2 = 0.05$.

\subsection{Comparison with Previous Approaches}
\label{sec:compare}

\begin{table}[!t]
    \caption{Comparison of video generation results. The best and second-best results are highlighted with \textbf{bold} and {\ul underline} formatting.}
    \label{tab:exp_video_compare}
    \resizebox{1\textwidth}{!}{
    \begin{tabular}{c|c|c||cc|cc||c}
    \toprule
    \multirow{2.5}{*}{\textbf{Method}} & \multirow{2.5}{*}{\textbf{Type}}   & \multirow{2.5}{*}{\textbf{Training}} & \multicolumn{2}{c|}{\textbf{Content Quality}} & \multicolumn{2}{c|}{\textbf{Dynamic Quality}}    & \multirow{2.5}{*}{\textbf{Overall $\uparrow$}} \\ \cmidrule{4-7}
                            &                         &                           & \textbf{Subj. $\uparrow$}         & \textbf{Bkg. $\uparrow$}     & \textbf{TSmoo. $\uparrow$} & \textbf{Dyn. $\uparrow$} &                            \\ \midrule
    Hunyuan-13B~\cite{kong2024hunyuanvideo}             & \multirow{2}{*}{3D V}   & \multirow{2}{*}{w/o}      & \textbf{0.9632}   & \textbf{0.9627}  & {\ul 0.9889}        & 0.4900           & 0.4493                     \\
    Wan-14B~\cite{jiang2025vace}                 &                         &                           & {\ul 0.9576}      & {\ul 0.9620}     & 0.9829              & 0.8476           & 0.7675                     \\ \midrule
    Animate Anyone~\cite{hu2024animate}          & \multirow{2}{*}{2.5D V} & \multirow{4}{*}{w/}       & 0.9206            & 0.9302           & 0.9671              & {\ul 0.9867}     & 0.8172                     \\
    Easy Animate~\cite{xu2024easyanimate}            &                         &                           & 0.9243            & 0.9358           & 0.9657              & \textbf{0.9933}  & 0.8297                     \\ \cmidrule{1-2} \cmidrule{4-8} 
    CogVideoX-5B~\cite{yang2025cogvideox}            & 3D V                    &                           & 0.9404            & 0.9477           & 0.9858              & \textbf{0.9933}  & {\ul 0.8727}                     \\
    Ours                    & 3D V\&M                 &                           & 0.9500            & 0.9533           & \textbf{0.9898}     & 0.9801           & \textbf{0.8785}                     \\ \bottomrule
    \end{tabular}
    }
    \end{table}

\begin{figure}[!t]
    \centering
    \includegraphics[width=\linewidth]{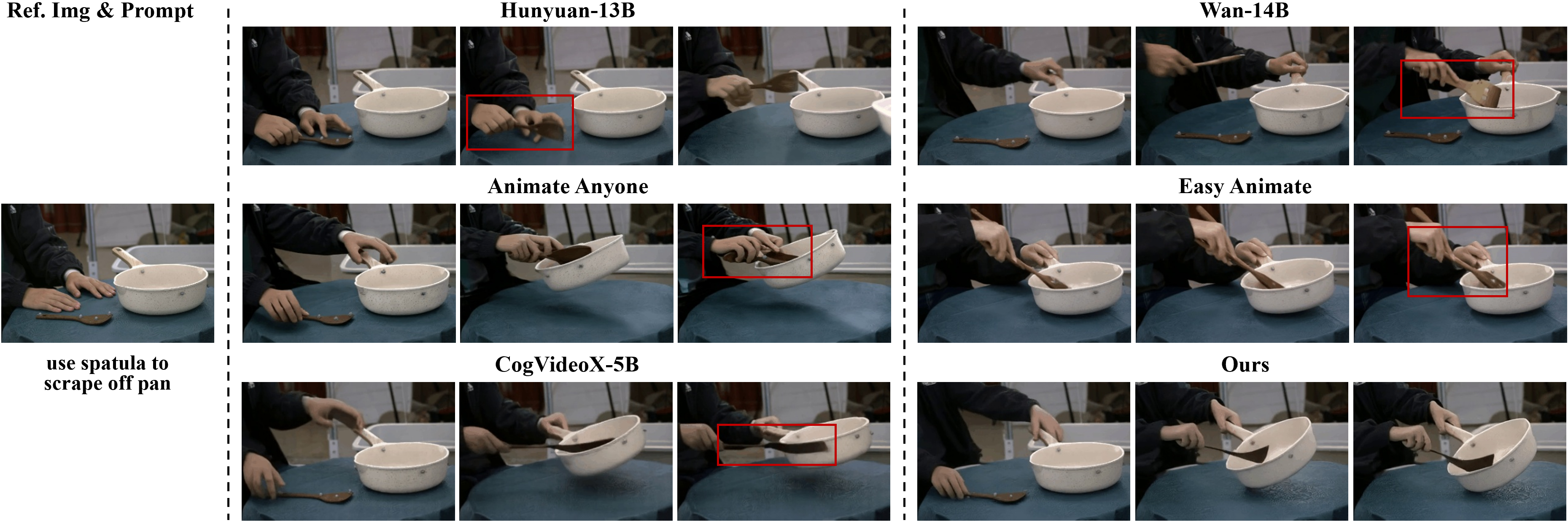}
    \caption{Visualization of videos. Red boxes highlight artifacts such as deformation, hallucinations, distortion, and implausible motions. 
    }
    \label{fig:comparison_video}
\end{figure}

\begin{figure}[!t]
    \centering
    \includegraphics[width=\linewidth]{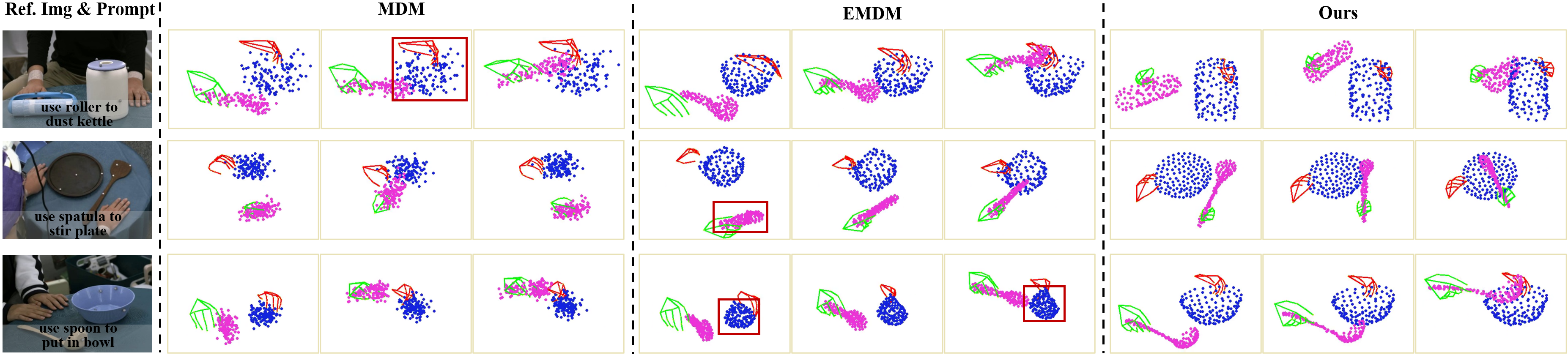}
    \caption{Quantitative comparison of motion generation results. Our method achieves smoother motions, sharper object point clouds, and higher instruction compliance.}
    \label{fig:comparison_motion}
\end{figure}

\textbf{Baselines.} For video generation performance, we compare with video models that following the image-animation paradigm (2.5D Video Models), including Animate Anyone~\cite{hu2024animate} and Easy Animate~\cite{xu2024easyanimate}, as well as native 3D large video models, including Hunyuan-13B~\cite{kong2024hunyuanvideo}, Wan-14B~\cite{jiang2025vace}, and CogVideoX-5B~\cite{yang2025cogvideox}. 
Particularly, due to the high training costs of the first two 3D models, we directly utilized them for zero-shot inference.
For 3D motion generation quality, we compare with the classic MDM~\cite{tevet2023human} and its latest improved version EMDM~\cite{zhou2024emdm}. To ensure fair comparison, we modified their code to align the experimental setup with ours.

\textbf{Quantitative and Qualitative Evaluation.} As shown in Table~\ref{tab:exp_video_compare}, our method achieves the highest overall score in video generation.
Notably, individual metrics often conflict: high scores in one aspect may compromise others. For instance, Hunyuan-13B~\cite{kong2024hunyuanvideo} attains top subject/background consistency and second-highest temporal smoothness, but its near-static outputs (dynamic degree: 0.49) yield the lowest overall score.
Wan-14B~\cite{jiang2025vace} also exhibits the same phenomenon, and the lack of TACO dataset fine-tuning results in poor instruction adherence and hallucinations (Fig.~\ref{fig:comparison_video}, Row 1).
The 2.5D image animation methods~\cite{hu2024animate, xu2024easyanimate} achieve high dynamic degree yet exhibit inadequate content consistency and temporal smoothness, visually manifesting as distortions and temporal flickering (Fig.~\ref{fig:comparison_video}, Row 2).
CogVideoX-5B~\cite{yang2025cogvideox} achieves the second-highest overall score, but the generated videos still exhibit inconsistencies, as shown on the left of the last row of Fig.~\ref{fig:comparison_video}.
In contrast, our method benefits from the synchronized modeling of visual and dynamic, resulting in better comprehensive performance.


\begin{wraptable}[9]{r}{0.5\linewidth}
\raggedleft
    \caption{Quantitative comparison against motion generation methods. The best results are in \textbf{bold}.}
    \label{tab:exp_motion_compare}
    \centering
    \resizebox{0.5\textwidth}{!}{
\begin{tabular}{c|cccc}
\toprule
\multirow{2.5}{*}{\textbf{Method}} & \multicolumn{2}{c}{\textbf{Hand}}     & \textbf{Object}             & \textbf{HOI}             \\ \cmidrule{2-5} 
                        & \textbf{MPJPE $\downarrow$}         & \textbf{MSmoo. $\downarrow$}    & \textbf{Cham. $\downarrow$} & \textbf{FID $\downarrow$}            \\ \midrule
MDM~\cite{tevet2023human}                     & 0.3382          & 0.0365          & 0.7915             & 0.4056          \\
EMDM~\cite{zhou2024emdm}                    & 0.3255          & 0.0306          & 0.7788             & 0.3681          \\
Ours                    & \textbf{0.1087} & \textbf{0.0255} & \textbf{0.1577}    & \textbf{0.1050} \\ \bottomrule
\end{tabular}
}
\end{wraptable}

For motion generation, our method achieves superior performance across all metrics, as shown in Tab.~\ref{tab:exp_motion_compare}. 
Qualitative results in Fig.~\ref{fig:comparison_motion} reveal that MDM~\cite{tevet2023human} and EMDM~\cite{zhou2024emdm} produce motions with poor instruction compliance and frame consistency.
This stems from two limitations: 1) They compress both reference images and text prompts into 512 dimensions through the CLIP encoder, then simply concatenate them as denoising conditions, which dilutes the instruction signal. 2) Their motion models lack vision awareness, causing large discrepancies between generated point clouds and references.
Contrastly, our approach not only preserves input condition effectiveness through a triple-modality adaptive modulation mechanism, but also enhances object point cloud consistency with low-level visual priors.

\begin{figure}[!t]
    \centering
    \includegraphics[width=0.3\linewidth]{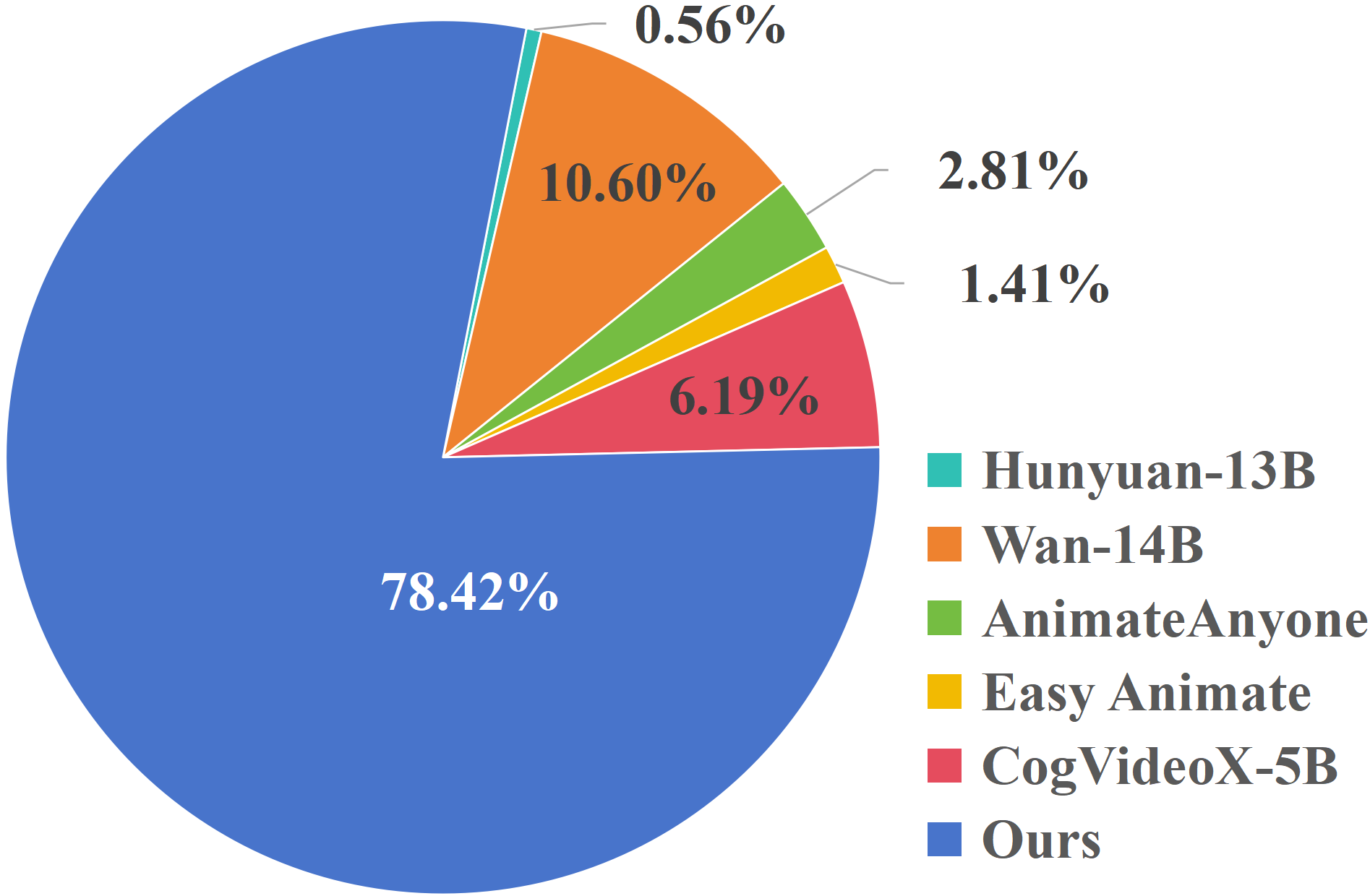}
    \hspace{1cm}
    \includegraphics[width=0.242\linewidth]{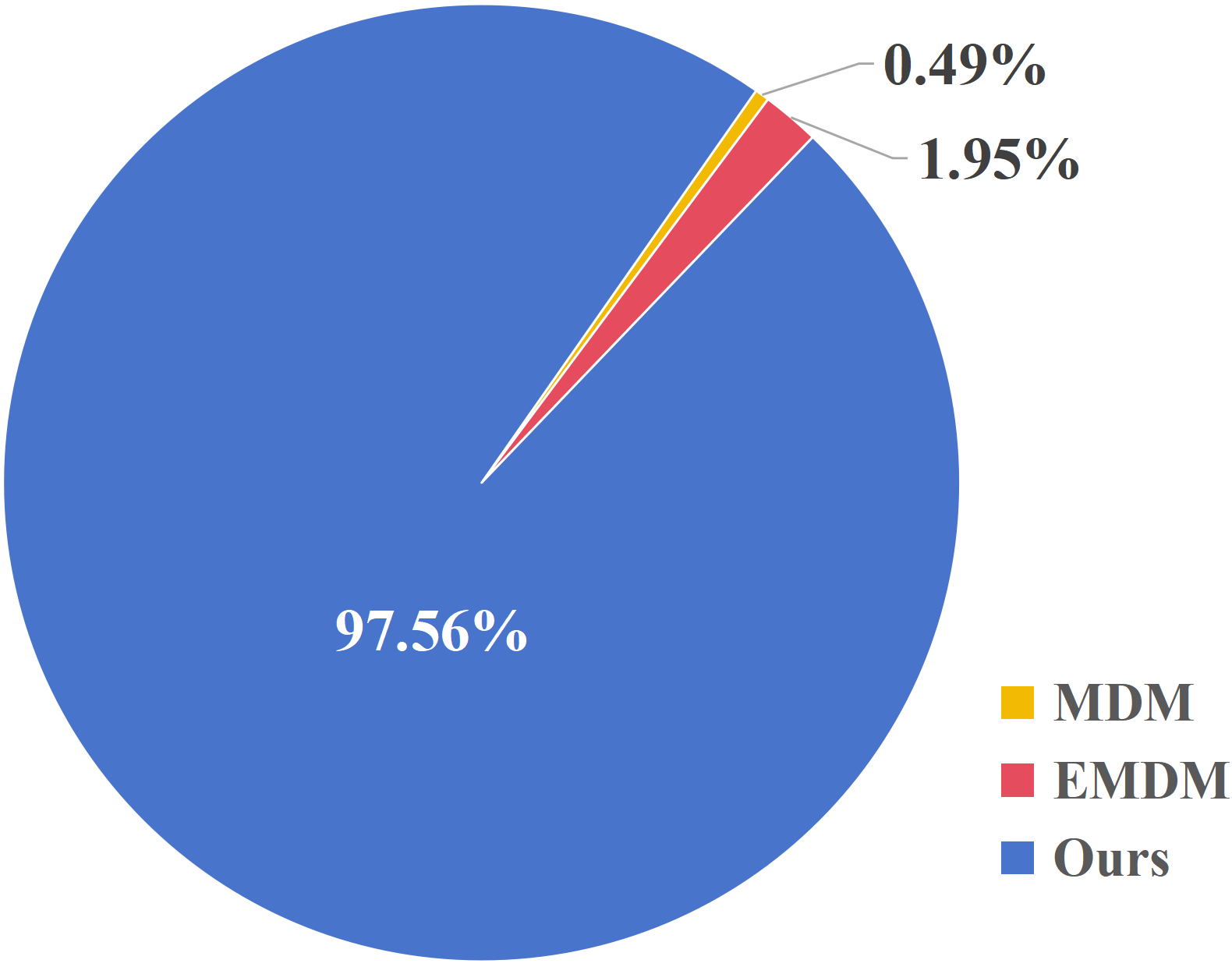} \\
    \flushleft \hspace{3.7cm} (a) \hspace{4.8cm} (b)
    \caption{User studies for generated videos (a) and motions (b). We received 1,066 and 410 valid responses respectively, and our method significantly outperforms other baselines.}
    \label{fig:user_study}
\end{figure}

\textbf{User Study.} To validate our method's effectiveness, we conducted user studies for video and motion generation (Fig.~\ref{fig:user_study}). For video generation, 26 image-prompt pairs were used to generate videos with six models each, yielding 1,066 valid responses from 41 participants. Our method achieved a 78.42\% preference rate, significantly outperforming all baselines. In motion generation, 10 image-prompt pairs produced 410 valid responses, with our results surpassing the baseline in 97.56\% of cases. These results demonstrate the clear advantages of our video-motion synchronous diffusion model.

\begin{figure}[!t]
    \centering
    \includegraphics[width=\linewidth]{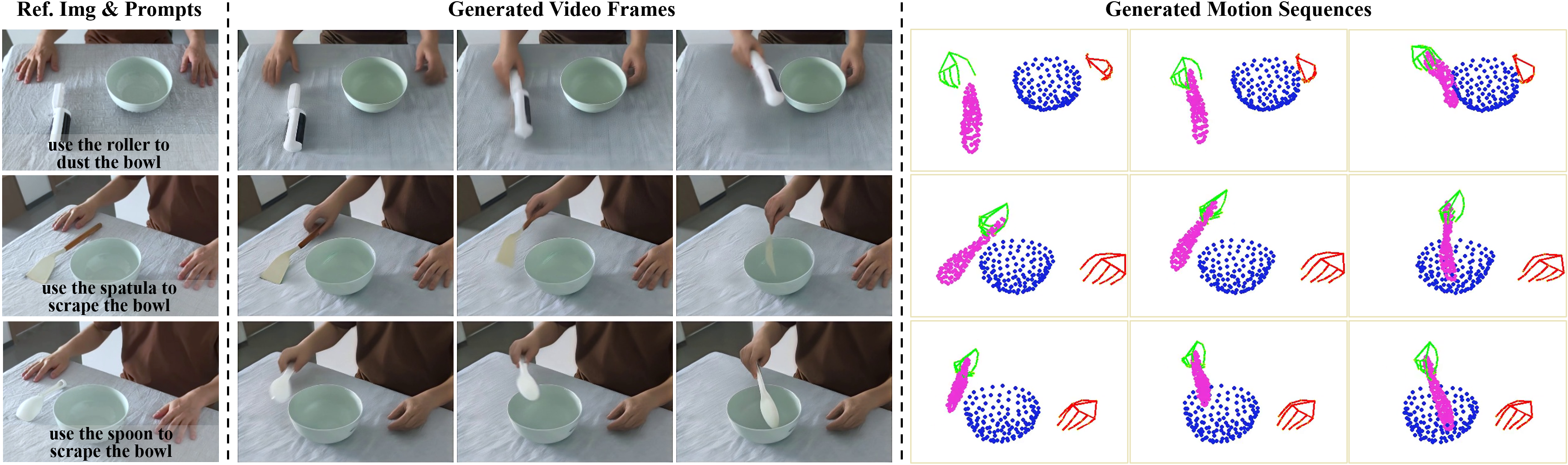}
    \caption{Zero-shot inference on real-world data. We collect data with everyday objects and generate high-fidelity, plausible HOI videos and 3D motions, demonstrating the generalizability of our method.
    }
    \label{fig:zeroshot}
\end{figure}

\subsection{Zero-shot Generalization on Real-world Data}
\label{sec:generalize}

We design manipulation tasks using common household objects such as rollers, spatulas, spoons, and bowls, and collect image-prompt pairs. These data are then input into our synchronized diffusion model to generate HOI videos and 3D interactions, as shown in Fig.~\ref{fig:zeroshot}. This demonstrates that our method can be easily generalized to real-world data.

\subsection{Ablation Study}
\label{sec:ablation}

\begin{table}[!t]
    \caption{Ablation studies on the synchronized diffusion and the vision-aware 3D interaction diffusion.}
    \label{tab:ablation}
    \centering
    \resizebox{1\textwidth}{!}{
    \begin{tabular}{c||cc|cc|c||cc|c|c}
    \toprule
    \multirow{2.5}{*}{\textbf{Varients}}       & \multicolumn{2}{c|}{\textbf{Content}} & \multicolumn{2}{c|}{\textbf{Dynamic}}    & \multirow{2.5}{*}{\textbf{Overall $\uparrow$}} & \multicolumn{2}{c|}{\textbf{Hand}} & \textbf{Object}             & \textbf{HOI}     \\ \cmidrule{2-5} \cmidrule{7-10} 
                                    & \textbf{Subj. $\uparrow$}  & \textbf{Bkg. $\uparrow$} & \textbf{TSmoo. $\uparrow$} & \textbf{Dyn. $\uparrow$}  &                            & \textbf{MPJPE $\downarrow$}   & \textbf{MSmoo. $\downarrow$}  & \textbf{Cham. $\downarrow$} & \textbf{FID $\downarrow$}    \\ \midrule
    SViMo w/ VID (Ours)            & {\ul{0.9534}}          & \textbf{0.9546}             & \textbf{0.9883}              & \textbf{0.9784}         & \textbf{0.8800}                     & \textbf{0.0121}    & \textbf{0.0053}        & \textbf{0.0019}             & \textbf{0.0100} \\
    SViMo w/ Inter. Guid.  & 0.9522          & \textbf{0.9546}             & 0.9877              & {\ul 0.9768}         & {\ul 0.8770}                     & 0.0157    & 0.0060        & 0.0022             & \textbf{0.0100} \\
    SViMo w/ Grad. Cons. & 0.9499          & 0.9525             & {\ul 0.9881}              & 0.9757         & 0.8723                     & {\ul 0.0141}    & {\ul 0.0058}        & {\ul 0.0021}             & {\ul 0.0124} \\
    SViMo w/o VID                  & \textbf{0.9543}          & {\ul 0.9545}             & \textbf{0.9883}              & 0.9686           & 0.8719                    & 0.0195    & 0.0070         & 0.0037             & 0.0546 \\
    VModel w/ Pred. Mot.  & 0.9356        & 0.9392             & 0.9858              & 0.9675          & 0.8381                     & 0.0202    & 0.0074        & 0.0040             & 0.0575  \\ \bottomrule
    \end{tabular}
    }
    \end{table}

\textbf{Effectiveness of Symbiotic Diffusion.} 
We argue that integrating visual priors and physical dynamics into a synchronized diffusion process is essential for HOI video and motion generation. 
To validate our synchronized diffusion mechanism: 
(1) We first remove VID to avoid confounding factors (\textbf{SViMo w/o VID}). 
(2) Then we decompose it into two independent components: a motion generation model with only motion loss and a video generation model conditioned on groundtruth motion (\textbf{VModel w/ GT Mot. Guid.}). 
After training these models independently, we use the predicted motions from the former as conditions for the latter during inference (\textbf{VModel w/ Pred. Mot.}). 
\begin{wrapfigure}[13]{r}{0.4\textwidth}
    \centering
    \includegraphics[width=0.9\linewidth]{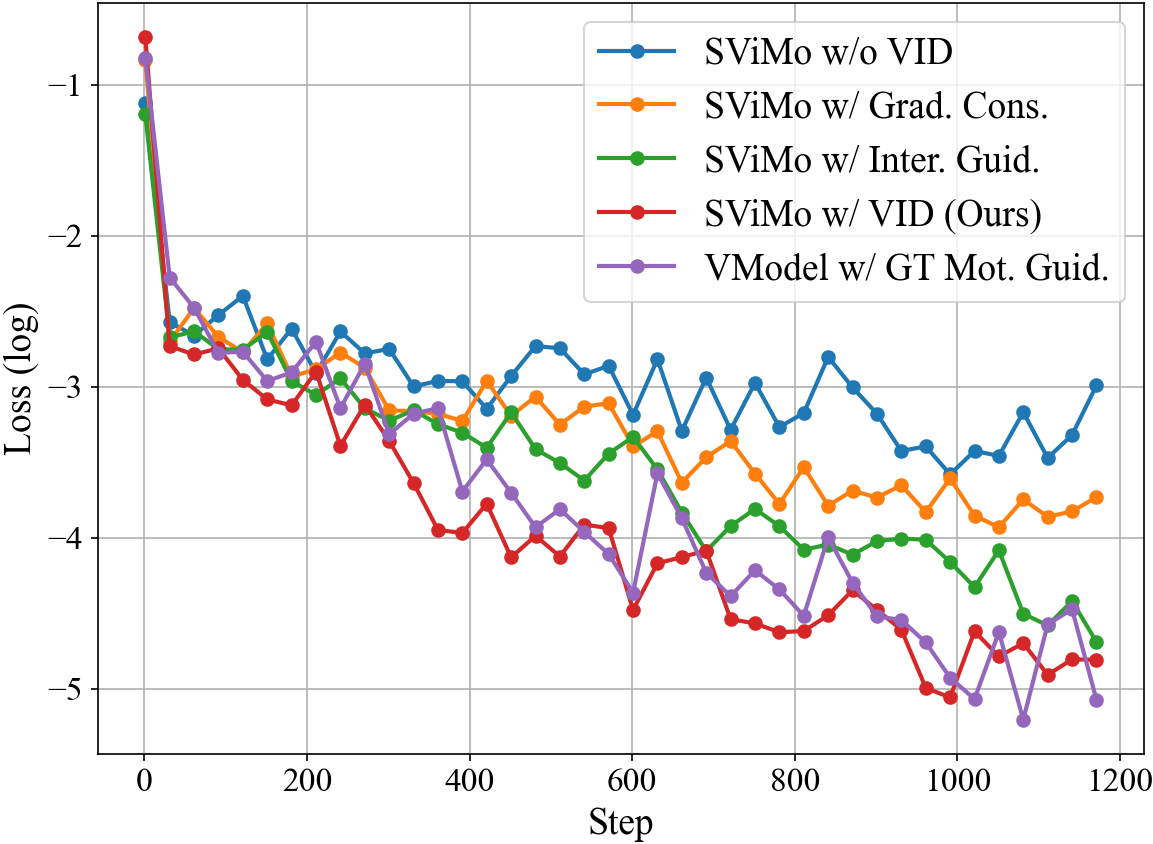}
    \caption{Video loss curves of different variants during the training process.}
    \label{fig:ablation_training_loss}
\end{wrapfigure} 
The last two rows of Tab.~\ref{tab:ablation} show that modeling video and motion independently not only leads to a 3.88\% decrease in video overall score (0.8719 \vs 0.8381), but also results in a 5.31\% degradation in motion FID (0.0546 \vs 0.0575).
This highlights the importance of our synchronous diffusion model in enabling feature-level synergy between video and motion.
This highlights the advantage of integrating visual priors and motion dynamics for our method.

\textbf{Impact of Vision-aware 3D Interaction Diffusion.} 
The vision-aware 3D interaction diffusion model forms a closed-loop feedback and co-evolution mechanism with the synchronized video-motion diffusion, by injecting interaction guidance and gradient constraints into the latter.
To validate its effectiveness, we conduct four variants: removing VID entirely (\textbf{SViMo w/o VID}), applying only gradient constraints (\textbf{SViMo w/ Grad. Cons.}), providing only interaction guidance (\textbf{SViMo w/ Inter. Guid.}), and preserving the full VID (\textbf{SViMo w/ VID}). 
Evaluation results in Tab.~\ref{tab:ablation} and training loss curves in Fig.~\ref{fig:ablation_training_loss} show that direct interaction guidance slightly outperforms gradient constraints, while the complete VID achieves the best performance.

\section{Conclusion}


Recognizing the synergistic co-evolution between the modeling of the visual appearance and that of the motion dynamics, we propose a synchronized diffusion model that simultaneously integrates hand-object interaction video generation and motion synthesis within a unified diffusion process. 
This approach produces both visually faithful and dynamically plausible results. Moreover, we introduce a vision-aware 3D interaction diffusion model that provides interaction guidance and gradient constraints to the synchronized denoising process of the above model, forming a closed-loop optimization pipeline that enhances video-motion consistency. 
Notably, our method requires no predefined conditions and demonstrates zero-shot generalization capabilities for real-world applications. 
This synchronized diffusion paradigm offers a promising pathway for fusing and aligning multimodal representations as well as building world models capable of understanding complex concepts. And we believe it has potential applicability across multiple domains.

\begin{ack}
    This work was supported by National Natural Science Foundation of China (NSFC) No.62125107 and No.62272172.
\end{ack}


{
\small
\bibliographystyle{plain}
\bibliography{myref}
}


\clearpage
\appendix

\section*{\centering SViMo: Synchronized Diffusion for Video and Motion Generation in Hand-object Interaction Scenarios (Appendix)}


In this supplementary material, we provide additional details on our methodology and experiments, and discuss the limitations, thereby enabling a more comprehensive and in-depth understanding of our proposed Synchronized diffusion for video and motion generation (SViMo). Below is the outline of all contents.



\hypersetup{linkbordercolor=black,linkcolor=black}

\setlength{\cftbeforesecskip}{0.5em}
\cftsetindents{section}{0em}{1.8em}
\cftsetindents{subsection}{1em}{2.5em}

\etoctoccontentsline{part}{Appendix}
\localtableofcontents
\hypersetup{linkbordercolor=red,linkcolor=red}

\section{Pseudo-code for the Training and Inference Phases}
\label{sec:pseudo-code}

In the main text, we introduced the Synchronized Video-Motion Diffusion $\mathcal{G}_\theta$ (SViMo, Sec. 3.3), which focuses on video generation. We also presented  the Vision-aware 3D Interaction Diffusion $\mathcal{M}_\phi$ (VID, Sec. 3.4), designed for motion generation. Additionally, we described a closed-loop feedback mechanism between these two components (Sec. 3.5). Next, we will provide a more detailed description of the training (Alg.~\ref{alg:joint_train}) and inference process (Alg.~\ref{alg:joint_inference}) using pseudocode.

\begin{algorithm}[H] 
\caption{Joint training process of SViMo and VID.}
\label{alg:joint_train}
\begin{algorithmic}[1]
\Require Reference image $\boldsymbol{I}$, text prompt $\boldsymbol{P}$, target video  $\boldsymbol{V}$, target 3D motion $(\boldsymbol{h}, \boldsymbol{o})$, frozen video VAE encoder $\mathcal{E}$, SViMo network $\mathcal{G}_\theta$, VID network $\mathcal{M}_\phi$.
\Ensure Optimized parameters of SViMo ($\theta^\star$) and VID ($\phi^\star$).
\State $\boldsymbol{M} = \text{Proj}(\boldsymbol{h}, \boldsymbol{o})$ \Comment{rendered motion video projection}
\State $\boldsymbol{z}_0^V = \mathcal{E}(\boldsymbol{V})$, $\boldsymbol{z}_0^M = \mathcal{E}(\boldsymbol{M})$, $\boldsymbol{z}_I = \mathcal{E}(\boldsymbol{I})$ \Comment{calculate latent codes}
\While {not converged}
    \State $t \sim \mathcal{U}\{1, \cdots, T\}$ \Comment{sample time step $t$}
    \State Calculate diffused data $\boldsymbol{z}_t^V$, $\boldsymbol{z}_t^M$, $\boldsymbol{h}_t$, $\boldsymbol{o}_t$ \Comment{following Eq. 1}
    \State $(\tilde{\boldsymbol{h}}_{0}, \tilde{\boldsymbol{o}}_{0}) = \mathcal{M}_\text{no-grad}\left((\boldsymbol{h}_t, \boldsymbol{o}_t), ({\boldsymbol{z}}_{t}^V, {\boldsymbol{z}}_{t}^M), t\right)$ \Comment{following Eq. 6}
    \State $\tilde{\boldsymbol{M}} = \text{Proj}(\tilde{\boldsymbol{h}}_0, \tilde{\boldsymbol{o}}_0)$, $\tilde{\boldsymbol{z}}_0^M = \mathcal{E}(\tilde{\boldsymbol{M}})$ \Comment{direct interaction guidance following Eq. 7}
    \State $(\hat{\boldsymbol{z}}_{0,\theta}^V, \hat{\boldsymbol{z}}_{0,\theta}^M) = \mathcal{G}_\theta(\boldsymbol{z}_t^V \oplus \boldsymbol{z}_I, \boldsymbol{z}_t^M \oplus \tilde{\boldsymbol{z}}_0^M, \boldsymbol{P}, t)$ \Comment{inverse denoising}
    \State $(\hat{\boldsymbol{h}}_{0, \phi}, \hat{\boldsymbol{o}}_{0, \phi}) = \mathcal{M}_\phi\left((\boldsymbol{h}_t, \boldsymbol{o}_t), (\hat{\boldsymbol{z}}_{0,\theta}^V, \hat{\boldsymbol{z}}_{0,\theta}^M), t\right)$ \Comment{indirect gradient constraint (Eq. 6,7)}
    \State $\mathcal{L} = \omega_1 \mathcal{L}_{\text{SViMo}} + \omega_2 \mathcal{L}_{\text{VID}}$ \Comment{calculate loss following Eq. 5,6}
    \State update parameters $\theta$ and $\phi$ by gradient descent
\EndWhile
\State \Return $\theta^\star = \theta$, $\phi^\star = \phi$
\end{algorithmic}
\end{algorithm}

\begin{algorithm}[H] 
\caption{Generation process of videos and motions.}
\label{alg:joint_inference}
\begin{algorithmic}[2]
\Require Reference image $\boldsymbol{I}$, text prompt $\boldsymbol{P}$, parameters of the noise scheduler $\{\alpha_t, \beta_t\}_{t=1}^{T}$, frozen video VAE encoder $\mathcal{E}$ and corresponding decoder $\mathcal{D}$, trained SViMo network $\mathcal{G}_\theta$ and VID network $\mathcal{M}_\phi$.
\Ensure Generated HOI video $\boldsymbol{V}$ and 3D motion $({\boldsymbol{h}}_0, {\boldsymbol{o}}_0)$.

\State $\boldsymbol{z}_I = \mathcal{E}(\boldsymbol{I})$ \Comment{calculate latent codes}
\State $\boldsymbol{z}_T^V \sim \mathcal{N}(\boldsymbol{0}, \boldsymbol{I}), \boldsymbol{z}_T^M \sim \mathcal{N}(\boldsymbol{0}, \boldsymbol{I}), (\boldsymbol{h}_T, \boldsymbol{o}_T) \sim \mathcal{N}(\boldsymbol{0}, \boldsymbol{I})$ \Comment{initialization}
\For{$t = T, \cdots, 1$}
    \State $(\tilde{\boldsymbol{h}}_{0}, \tilde{\boldsymbol{o}}_{0}) = \mathcal{M}_\text{no-grad}\left((\boldsymbol{h}_t, \boldsymbol{o}_t), ({\boldsymbol{z}}_{t}^V, {\boldsymbol{z}}_{t}^M), t\right)$ \Comment{following Eq. 6}
    \State $\tilde{\boldsymbol{M}} = \text{Proj}(\tilde{\boldsymbol{h}}_0, \tilde{\boldsymbol{o}}_0)$, $\tilde{\boldsymbol{z}}_0^M = \mathcal{E}(\tilde{\boldsymbol{M}})$ \Comment{direct interaction guidance following Eq. 7}
    \State $(\hat{\boldsymbol{z}}_{0}^V, \hat{\boldsymbol{z}}_{0}^M) = \mathcal{G}_\theta(\boldsymbol{z}_t^V \oplus \boldsymbol{z}_I, \boldsymbol{z}_t^M \oplus \tilde{\boldsymbol{z}}_0^M, \boldsymbol{P}, t)$ \Comment{inverse denoising}
    \State $(\hat{\boldsymbol{h}}_{0}, \hat{\boldsymbol{o}}_{0}) = \mathcal{M}_\phi\left((\boldsymbol{h}_t, \boldsymbol{o}_t), (\hat{\boldsymbol{z}}_{0}^V, \hat{\boldsymbol{z}}_{0}^M), t\right)$ \Comment{indirect gradient constraint (Eq. 6,7)}
    \State $\boldsymbol{\mu}_{V} = \left(\frac{\sqrt{\alpha_t} (1-\bar{\alpha}_{t-1})}{1-\bar{\alpha}_{t}} \boldsymbol{z}_t^V + \frac{\sqrt{\bar{\alpha}_{t-1}} (1-\alpha_t)}{1-\bar{\alpha}_{t}} \hat{\boldsymbol{z}}_{0}^V\right)$, same as $\boldsymbol{\mu}_{M}$ and $\boldsymbol{\mu}_{h,o}$
    \State $\sigma^2 = \frac{1 - \bar{\alpha}_{t-1}}{1 - \bar{\alpha}_t} (1 - \alpha_t)$ \Comment{constant value}
    \State $\boldsymbol{z}_{t-1}^V \sim \mathcal{N}\left(\boldsymbol{\mu}_V, \sigma^2 \boldsymbol{I}\right)$, $\boldsymbol{z}_{t-1}^M \sim \mathcal{N}\left(\boldsymbol{\mu}_M, \sigma^2 \boldsymbol{I}\right)$, $(\boldsymbol{h}_{t-1}, \boldsymbol{o}_{t-1}) \sim \mathcal{N}\left(\boldsymbol{\mu}_{h,o}, \sigma^2 \boldsymbol{I}\right)$
\EndFor
\State $\boldsymbol{V}=\mathcal{D}(\boldsymbol{z}_0^V)$ \Comment{decode into raw video}
\State \Return $\boldsymbol{V}$, $(\boldsymbol{h}_0, \boldsymbol{o}_0)$
\end{algorithmic}
\end{algorithm}


\section{More Implementation Details}
\label{sec:more_imp_details}


\subsection{VBench Metrics}

We use some of the evaluation metrics from VBench~\cite{huang2024vbench} for quantitative analysis of the generated videos. The details of its implementation could not be fully presented in Sec. 4.1 of the main text due to page constraints, so we provide a supplement here.

\textbf{Subject Consistency} score is used to evaluate whether a subject (\eg, a person or an object) remains consistent throughout a video. Specifically, we first extract image features from each frame of the video using DINO~\cite{caron2021emerging}. Then we calculate the cosine similarity between features of consecutive frames and between each frame and the first frame to characterize subject consistency. The calculation formula is as follows:
\begin{equation}
    \label{eq:subject_consistency}
    S_\text{Subj.}=\frac{1}{T-1}\sum_{t=2}^T\frac{1}{2}\left(\langle d_1\cdot d_t\rangle+\langle d_{t-1}\cdot d_t\rangle\right),
\end{equation}
where $d_t$ is the DINO  feature of the $t^{th}$ video frame.

\textbf{Background Consistency.} High-quality videos should not only ensure the subject's appearance remains consistent throughout the video, but also should not overlook the consistency of the background, such as scenes, rooms, and tabletops. VBench~\cite{huang2024vbench} computes the background consistency score by utilizing the cosine similarity of CLIP~\cite{radford2021learning} image features, with the calculation process being similar to Eq.~\ref{eq:subject_consistency}:
\begin{equation}
    \label{eq:bkg_consistency}
    S_\text{Bkg.}=\frac{1}{T-1}\sum_{t=2}^T\frac{1}{2}\left(\langle c_1\cdot c_t\rangle+\langle c_{t-1}\cdot c_t\rangle\right),
\end{equation}
where $c_t$ is the CLIP  feature of the $t^{th}$ video frame.

\textbf{Temporal Smoothness} score is based on prior knowledge that video motion should be smooth, \ie, linear or quadratic, over very short time intervals (a few consecutive video frames). To quantify the smoothness of a generated video $\boldsymbol{V} = \{v_1, v_2, \cdots, v_{2T}\}$, VBench~\cite{huang2024vbench} first removes odd-numbered frames $(v_1, v_3, \cdots, v_{2T-1})$. Then it uses a video frame interpolation model~\cite{li2023amt} to generate the removed odd-numbered frame sequence $(\hat{v}_1, \hat{v}_3, \cdots, \hat{v}_{2T-1})$. Next, it calculates the mean absolute error between the reconstructed frames and the original removed frames to measure whether the generated video satisfies the prior knowledge embedded in the video frame interpolation model. Finally, this value is normalized to the range [0, 1]:
\begin{equation}
    \label{eq:tsmoothness}
    S_\text{TSmoo.}=\frac{1}{T}\sum_{t=1}^T\frac{255 - \text{MAE}(v_{2t-1}, \hat{v}_{2t-1})}{255}.
\end{equation}

\textbf{Dynamic Degree.} The three metrics mentioned above, \ie, Subject Consistency, Background Consistency, and Temporal Smoothness, can individually quantify the performance of generated videos in certain specific aspects. However, a completely static video may achieve very high scores in them, which is not the desired outcome. 
To evaluate the dynamic degree of generated videos, VBench~\cite{huang2024vbench} employs RAFT~\cite{teed2020raft} to calculate optical flow strengths between adjacent frames. The average of the top 5\% highest optical flow intensities is taken as the video's dynamic score. Once this score exceeds a predefined threshold $\tau_{op}$, the video is considered dynamic, otherwise, it is deemed static. The overall dynamic score is calculated as follows:
\begin{equation}
    \label{eq:dynamic_degree}
    S_\text{Dyn.}=\frac{1}{N}\sum_{n=1}^N \mathbb{I}\left(\text{Avg}\left(\text{Top}_{t=2:T}^{5\%} \left(\text{RAFT}\langle v_{t-1}, v_t \rangle \right)\right) > \tau_{op} \right),
\end{equation}
where $N$ is the number of generated videos.

\subsection{Training Loss for 3D Motions}

We generate 3D motions using a vision-aware 3D interaction diffusion model. During training, we apply separate supervision for hand and object motions. The hand loss is computed as:
\begin{equation}
    \label{eq:hand_loss}
    \mathcal{L}_\text{hand} =  \left\| \boldsymbol{h} - \hat{\boldsymbol{h}} \right\|_2^2 + 0.2 \cdot \left\| \boldsymbol{h}^\prime - \hat{\boldsymbol{h}}^\prime \right\|_2^2 + 0.05 \cdot \left\| \boldsymbol{h}^{\prime \prime} - \hat{\boldsymbol{h}}^{\prime \prime} \right\|_2^2,
\end{equation}
where the three terms represent the hand pose and its first-order and second-order differences between frames.
For the 3D object point cloud sequence $\boldsymbol{o} \in \mathbb{R}^{[B, N, K, 3]}$, we first split the object into tool $\boldsymbol{o}_\text{tool} \in \mathbb{R}^{[B, N, K/2, 3]}$ and target $\boldsymbol{o}_\text{target} \in \mathbb{R}^{[B, N, K/2, 3]}$, then calculate losses separately for these two components and average them to obtain the final object loss.
Firstly, we directly compute the Chamfer distance between each frame's generated point cloud and the ground truth to ensure consistency in shape contours. 
Furthermore, to enforce smoothness of the inter-frame motion, we first reshape the data into $[B \times N, K/2, 3]$, then calculate pairwise Chamfer distances between consecutive frames for both ground truth and generated motion, which captures the movement dynamic in the point cloud sequence. Finally, we compute the mean absolute error (MAE) between predicted values and ground truth motion dynamics:
\begin{equation}
    \label{eq:object_loss}
    \begin{aligned}
        \mathcal{L}_1 &=  \text{Avg}\left\{D_\text{chamf}\left( \boldsymbol{o}, \hat{\boldsymbol{o}}\right)\right\}, \\
        \mathcal{L}_\text{2} &=  \text{Avg}\left\{ \frac{1}{N-1} \sum_{n=2}^{N} \text{MAE} \left( D_\text{chamf}( \boldsymbol{o}_n, \boldsymbol{o}_{n-1}), D_\text{chamf}( \hat{\boldsymbol{o}}_n, \hat{\boldsymbol{o}}_{n-1})\right)
        \right\}, \\
        \mathcal{L}_\text{obj} &=  \mathcal{L}_1 + 0.1 \cdot \mathcal{L}_2,
    \end{aligned}
\end{equation}
where $\boldsymbol{o}$ consists of $\boldsymbol{o}_\text{tool}$ and $\boldsymbol{o}_\text{target}$.

\subsection{Pretrained 3D Interaction Reconstruction Model}

To assess the difference between the distribution of the generated 3D interactions and that of the ground truth motions, we trained a 3D action reconstruction model on the TACO~\cite{liu2024taco} dataset using the variational autoencoder architecture from Diverse Sampling~\cite{dang2022diverse}. During evaluation, we utilized features from the residual graph convolutional layers of this model to compute the FID score.


\section{Additional Experimential Results}

\subsection{More Ablation Studies}

Our model is built upon the CogVideoX-5B~\cite{yang2025cogvideox} model and extended to a video-motion joint generation model. Additional parameters include the input and output projection layers, and triple modality modulation modules in all 42 DiT Blocks, resulting in a total of 6.02 billion parameters. 
To reduce training computational consumption, we tested two model variants. First, we added motion modality modulation modules only to even-numbered DiT Blocks while retaining only the original text and video modulation modules in odd-numbered DiT Blocks, \ie, \textbf{Only Even Layers}. 
This design slightly reduced the total parameter count. 
Second, we applied \textbf{LoRA Training}, which uses low-rank decomposition of the original model parameters to significantly decrease the number of trainable parameters. Experimental results in Tab.~\ref{tab:ablation_lora_and_even_insert} show that both variants produce lower-quality videos and motion compared to our default settings. This indicates that effective alignment and fusion of video and motion modality features are essential for generating high-quality outputs.


\begin{table}[H]
    \caption{Ablation studies on the synchronized diffusion and the vision-aware 3D interaction diffusion.}
    \label{tab:ablation_lora_and_even_insert}
    \centering
    \resizebox{1\textwidth}{!}{
    \begin{tabular}{c||cc|cc|c||cc|c|c}
    \toprule
    \multirow{2.5}{*}{\textbf{Varients}}       & \multicolumn{2}{c|}{\textbf{Content}} & \multicolumn{2}{c|}{\textbf{Dynamic}}    & \multirow{2.5}{*}{\textbf{Overall $\uparrow$}} & \multicolumn{2}{c|}{\textbf{Hand}} & \textbf{Object}             & \textbf{HOI}     \\ \cmidrule{2-5} \cmidrule{7-10} 
                                    & \textbf{Subj. $\uparrow$}  & \textbf{Bkg. $\uparrow$} & \textbf{TSmoo. $\uparrow$} & \textbf{Dyn. $\uparrow$}  &                            & \textbf{MPJPE $\downarrow$}   & \textbf{MSmoo. $\downarrow$}  & \textbf{Cham. $\downarrow$} & \textbf{FID $\downarrow$}    \\ \midrule
    Ours            & {\ul{0.9534}}          & {\ul {0.9546}}             & {0.9883}              & \textbf{0.9784}         & \textbf{0.8800}                     & \textbf{0.0121}    & \textbf{0.0053}        & \textbf{0.0019}             & \textbf{0.0100} \\
    Only Even Layers & {\textbf{0.9535}}          & \textbf{0.9552}             & {\ul{0.9884}}              & {\ul{0.9757} }        & {\ul{0.8783}}                   & {\ul{0.0297}}    & {\ul{0.0103}}        & {\ul{0.0127}}             & {\ul{0.1368}} \\
    LoRA Training  & {0.9458}          & {0.9523}             & \textbf{0.9892}              & {0.9688}         & {0.8631}                     & {0.0671}    & {0.0211}        & {0.0555}             & {0.4671} \\ \bottomrule
    \end{tabular}
    }
    \end{table}

\subsection{Qualitative Results of Our SViMo}

\begin{figure}[H]
    \centering
    \includegraphics[width=\linewidth]{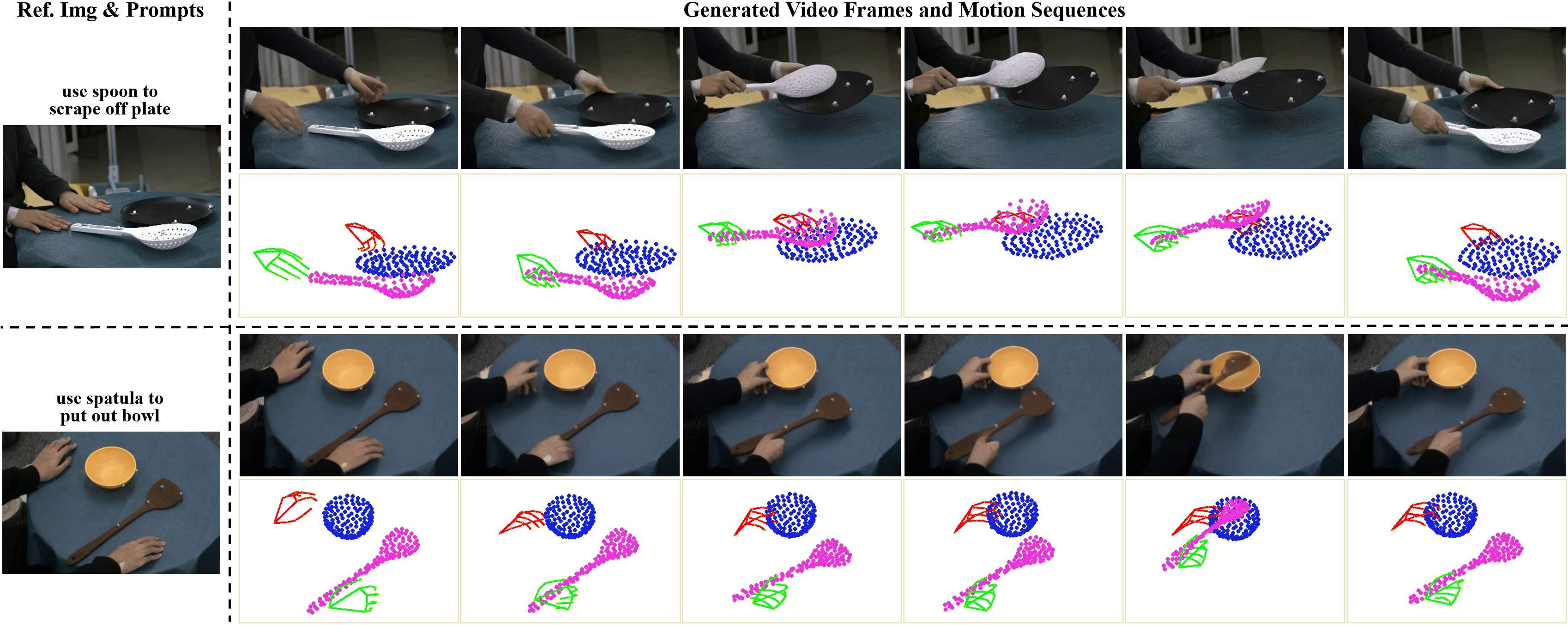}
    \caption{Visualization of the generated videos and motions. We provide vivid demonstrations in the video of the supplementary materials.
    }
    \label{fig:app_ours}
\end{figure}

We present two additional video and motion generation results as shown in Figure~\ref{fig:app_ours}. These results demonstrate videos and motions that are both reasonable and consistent. Considering that image frames cannot effectively showcase video effects, we provide videos in the supplementary materials for a more vivid demonstration.

\subsection{Comparisons of Our SViMo and Other Methods}

\begin{figure}[!t]
    \centering
    \includegraphics[width=\linewidth]{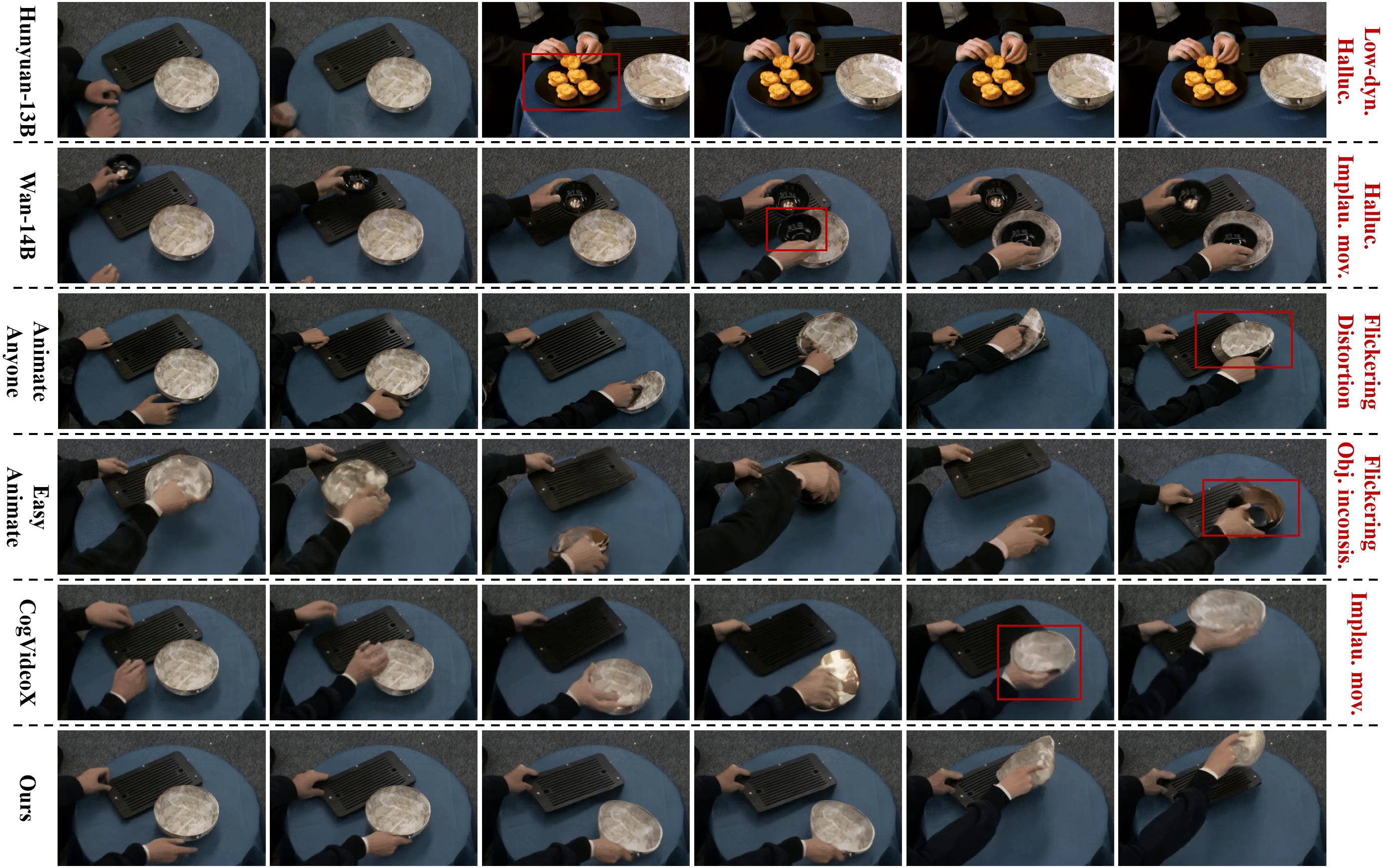}
    \caption{
        Comparison of video generation results. The artifacts in the videos generated by the baseline methods are highlighted in \textcolor{red}{red}. Refer to the video in the supplementary material for vivid demonstrations.
    }
    \label{fig:app_comp1}
\end{figure}

\begin{figure}[!t]
    \centering
    \includegraphics[width=\linewidth]{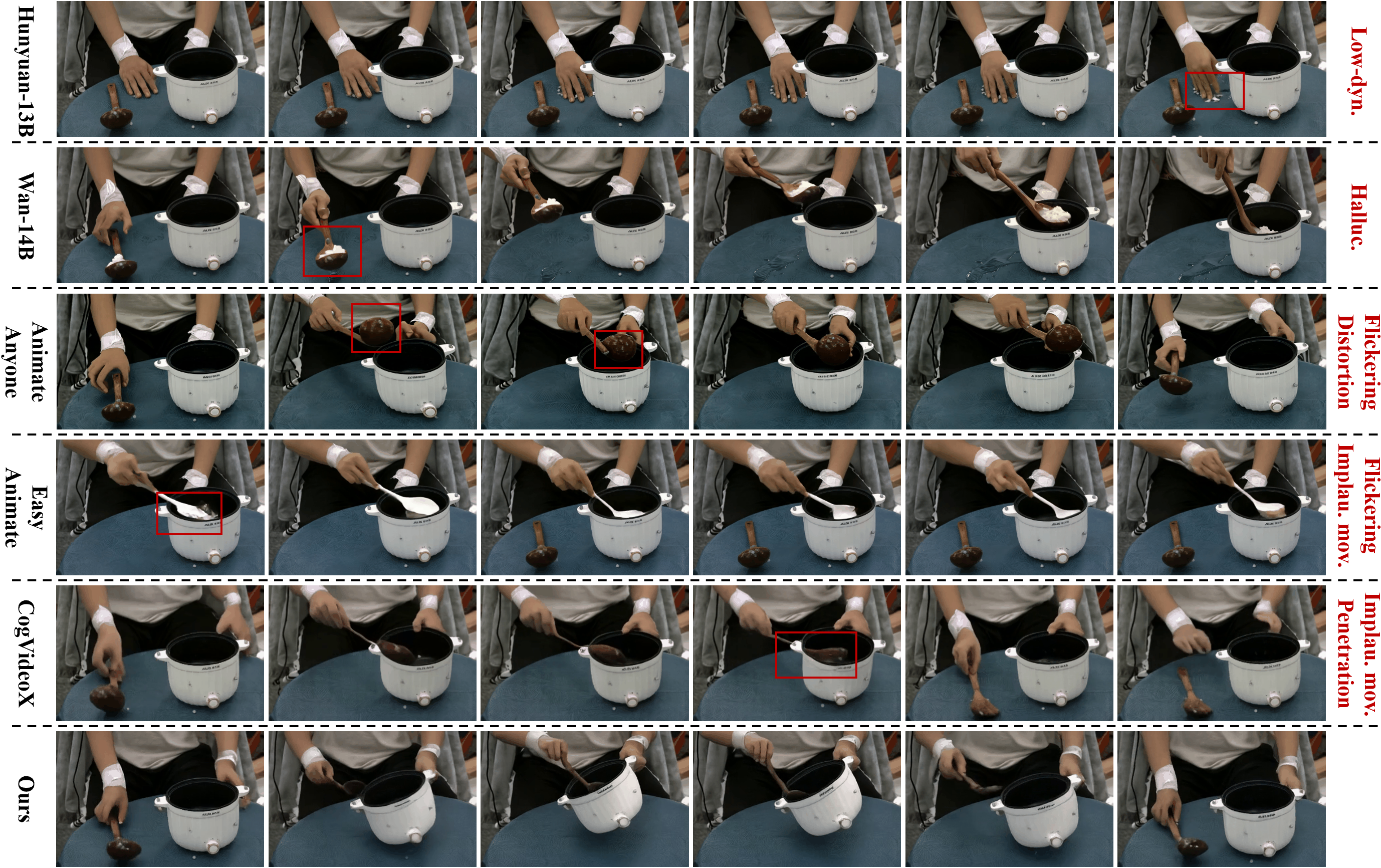}
    \caption{Comparison of video generation results. The artifacts in the videos generated by the baseline methods are highlighted in \textcolor{red}{red}. Refer to the video in the supplementary material for vivid demonstrations.
    }
    \label{fig:app_comp2}
\end{figure}

\begin{figure}[!t]
    \centering
    \includegraphics[width=\linewidth]{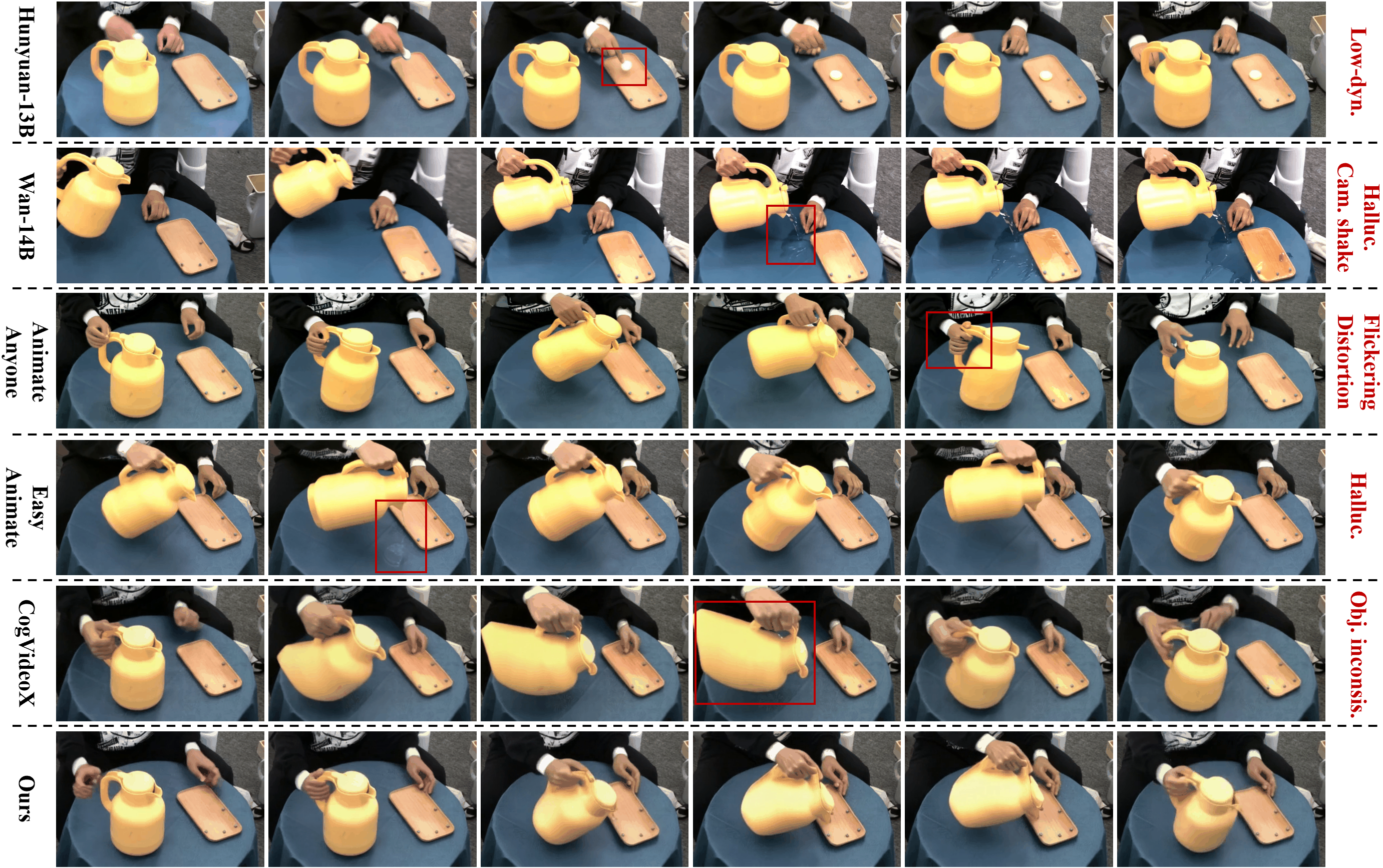}
    \caption{Comparison of video generation results. The artifacts in the videos generated by the baseline methods are highlighted in \textcolor{red}{red}. Refer to the video in the supplementary material for vivid demonstrations.
    }
    \label{fig:app_comp3}
\end{figure}

We present additional video generation results from three more cases, as shown in Figures~\ref{fig:app_comp1}, ~\ref{fig:app_comp2}, and ~\ref{fig:app_comp3}. 
Other baseline methods' generated results exhibit certain artifacts, such as flickering, distortion, and implausible movements, which are highlighted in red.
Our method demonstrates superior performance compared to theirs.

\subsection{Failure Cases}

\begin{figure}[!t]
    \centering
    \includegraphics[width=\linewidth]{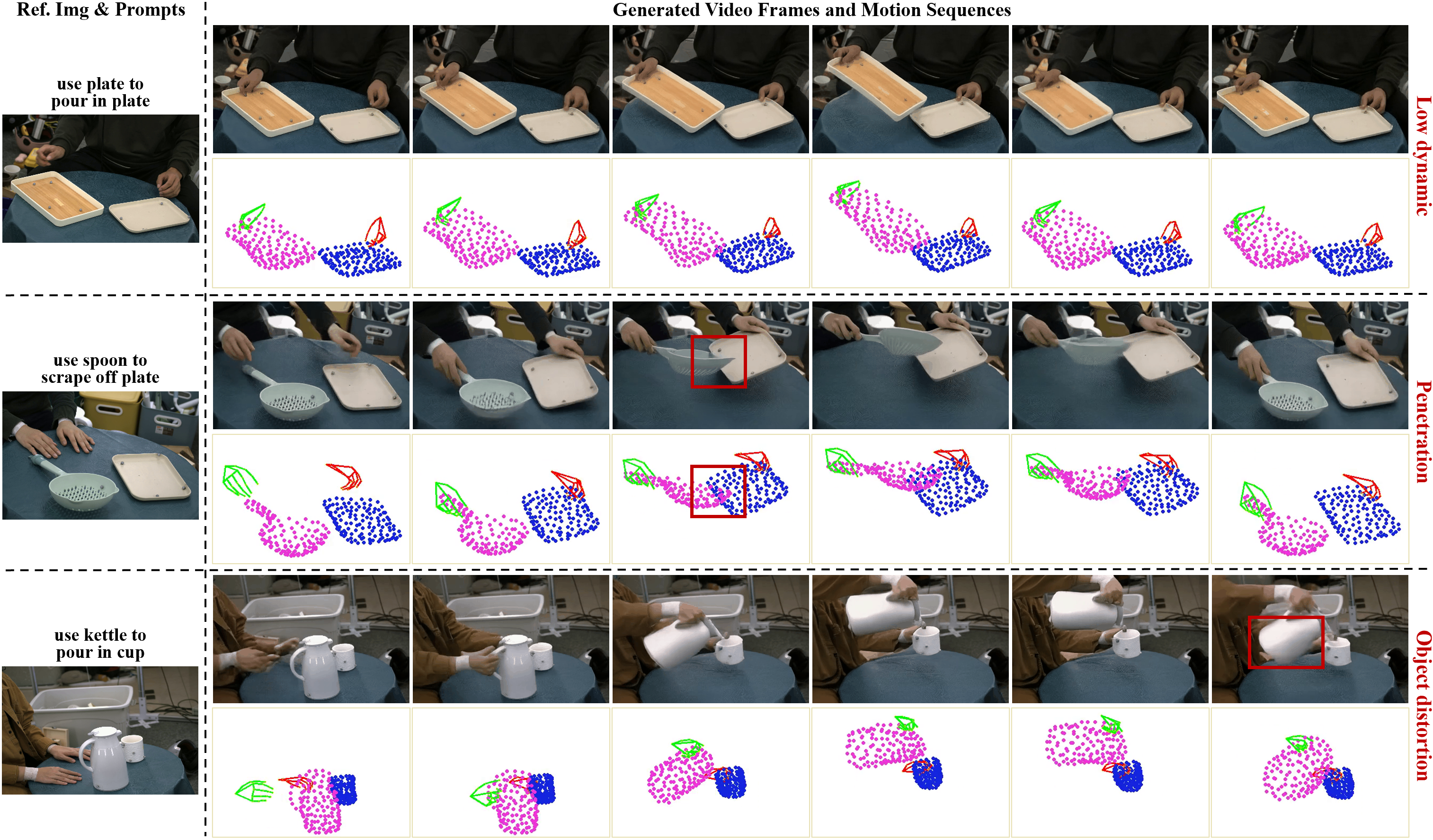}
    \caption{Visualization of failure cases. \textcolor{red}{Red} boxes highlight artifacts.
    }
    \label{fig:app_fail}
\end{figure}

While our approach can generate high-realistic videos and plausible motions in most cases, it occasionally exhibits certain artifacts such as penetration, low dynamics, and object inconsistency, as shown in Figure~\ref{fig:app_fail}.
However, the overall performance is still better than the baseline method. 
Addressing these issues, we need to explore stronger foundational models on one hand and leverage larger-scale video-motion datasets for training on the other hand.

\section{Limitations and Future Work}

In fine-grained hand-object interaction scenarios, simultaneously generating visually high-fidelity videos and physically plausible motions remains a challenging task. While our synchronized diffusion model has made significant progress in addressing this challenge, three key limitations still exist. 
First, our method relies on a foundation model trained on large-scale video data, which is then fine-tuned on a smaller dataset of video-3D motion pairs. 
Although the latter has a relatively smaller data scale, it remains essential for the expansion. 
Second, while our approach can generate diverse interaction motions, the produced 3D object point clouds are currently restricted to rigid, simple objects and struggle with structurally complex geometries. 
Third, the capabilities of the pre-trained foundation model directly impact both training efficiency and final performance. For instance, using lightweight LoRA (low-rank adaptation) strategies with the CogVideoX~\cite{yang2025cogvideox} foundation model results in suboptimal outcomes, even a full-parameter finetuned model has potential blurring artifacts when sampling at reduced resolution.

To address these challenges, future work should focus on three directions. 
First, replacing the non-differentiable 3D trajectory representation with differentiable neural representations (\eg, NeRF-style formulations~\cite{mildenhall2021nerf}) could enable video-only supervision without requiring explicit 3D annotations. 
This would transform our vision-aware 3D interaction diffusion model (VID) into a large reconstruction model, potentially resolving the second limitation. 
Second, continuous following of advanced open-source foundation models is necessary, as their evolving capabilities directly affect training stability and output quality. Third, integrating visual reinforcement learning strategies~\cite{liu2025visual} could further enhance generation fidelity. 
These improvements would collectively advance the field toward more robust and scalable hand-object interaction synthesis.






\section{Broader Impacts}

Our proposed synchronized video-motion diffusion model generates hand-object interaction videos alongside corresponding motion sequences, achieving both visual realism and physically plausible dynamics. This overcomes the limitations of prior approaches that prioritized only visual quality or physical accuracy in isolation. The model demonstrates generalization capabilities to real-world scenarios, making significant progress in this research domain.
Moreover, the joint diffusion paradigm offers valuable insights for cross-modal information integration in multimodal generative models. Practical applications span multiple fields, including game development, digital human animation, and embodied robotics, with potential societal benefits across these domains.

While acknowledging the positive impacts of this technology, we also recognize its dual-use nature. The system could be misused to generate synthetic media that violates privacy or perpetuates biases. To address these concerns, we commit to ethical development practices and transparent implementation protocols. Our research emphasizes proactive risk mitigation to ensure responsible innovation that maximizes social benefits while minimizing potential harms.


\end{document}